\documentclass[acmsmall,screen,nonacm]{acmart}

\setcopyright{none}                         
\renewcommand\footnotetextcopyrightpermission[1]{}   

\usepackage{graphicx}
\usepackage{placeins}
\usepackage{adjustbox}
\usepackage{booktabs}
\usepackage{capt-of}
\usepackage{amsmath}   
\usepackage[table]{xcolor}
\usepackage{enumitem}
\usepackage{fvextra}
\usepackage{DejaVuSans}
\usepackage{tikz}      
\usetikzlibrary{arrows.meta,shapes.geometric}
\usepackage{bookmark}
\usepackage[subdued,italic]{mathastext}
\MTDeclareVersion[sl]{figuremath}{T1}{DejaVuSans-TLF}{m}{n}
\newcommand{\figuremathfont}{\mathversion{figuremath}\sffamily}
\newcommand{\figuretitlefont}{\figuremathfont\bfseries\fontsize{9}{11}\selectfont}
\newcommand{\figurebodyfont}{\figuremathfont\fontsize{8}{10}\selectfont}
\newcommand{\figurenotefont}{\figuremathfont\fontsize{7}{9}\selectfont}
\definecolor{reportBlue}{HTML}{2F69AA}
\definecolor{reportOrange}{HTML}{E07B27}
\definecolor{reportGreen}{HTML}{2E8B57}
\definecolor{reportPurple}{HTML}{7450A0}
\definecolor{reportRed}{HTML}{C74646}
\usetikzlibrary{patterns}
\tikzset{
  report positive fill/.style={postaction={pattern=north east lines,pattern color=black!40}},
  report intermediate fill/.style={postaction={pattern=dots,pattern color=black!45}},
  report negative fill/.style={postaction={pattern=north west lines,pattern color=black!40}},
  every picture/.append style={line width=0.7pt},
  thick/.style={line width=1.2pt},
  very thin/.style={line width=0.4pt},
  report figure/.style={font=\figurebodyfont},
  report title/.style={font=\figuretitlefont},
  report note/.style={font=\figurenotefont,text=black!65},
  report line/.style={line width=0.7pt},
  report accent/.style={line width=1.2pt},
}

\newcommand{\researchpart}[2]{%
  \par\addpenalty{-300}
  \bookmarksetupnext{startatroot}%
  \phantomsection
  \addcontentsline{toc}{part}{Part #1: #2}%
  \par\nointerlineskip\vskip1.5\baselineskip
  \noindent\vbox{%
    \hsize=\linewidth
    {\LARGE\bfseries Part #1: #2\par}%
  }%
  \par\nobreak\kern1.25ex
}

\title[Decision Support Across the Venture Lifecycle]{From Ideas to Actions:\texorpdfstring{\\}{ }
       A Public-Data Decision-Support Toolchain\texorpdfstring{\\}{ }
       Across the Venture Lifecycle}
\author{Lei Qu}
\affiliation{%
  \institution{Shanghai Xing Yun Zhi Li AI Institute}
  \city{Shanghai}
  \country{China}
}
\email{mr.leiqu@gmail.com}
\date{September 15, 2026}

\begin{document}

\begin{abstract}
Founders face two linked decisions across the venture lifecycle: before founding,
whether an idea warrants pursuit; after founding, which operating actions and
capital partners fit the company's intended trajectory. The necessary evidence is
fragmented and often confined to subscription-only venture databases. We present
a two-part decision-assistance toolchain built from freely available public information.
Part~One combines structured proposal profiling, time-bounded market and moat
checks, and deterministic aggregation. Part~Two reconstructs auditable
investor--company event chains to analyze operating-action patterns and
post-investment behavior. We evaluate the proposal pipeline against historical
outcomes and apply the toolchain in AI-inference and chip-company studies,
yielding findings for both decision stages.

\textbf{Pre-founding.} \textbf{(a)} After threshold selection on a 198-company
development draw, the frozen system yields $F_{0.5}=0.5357$ [0.412, 0.655] on
an independently drawn, row-disjoint 198-company validation sample; the combined
396-row benchmark estimate is 0.6301. A paired Raw LLM baseline scores 0.2734 on
those 396 rows. In a separately drawn scale cohort, post-stratifying all 1{,}027
completed cases to the fixed execution-split composition yields 0.6506 [0.598,
0.707], with a 377-row composition-matched check yielding 0.6573. Together,
these checks strengthen the evidence that performance is not confined to the
original benchmark, although the incomplete scale run precludes a
deployment-grade generalization claim.
\textbf{(b)} The AI-inference market study identifies two distinct paths:
distribution-layer businesses offer the most replicable route to independent
profitability but face a limited revenue ceiling, whereas frontier-model ownership
offers greater capital-market upside at exceptional capital cost.
\textbf{Post-founding.} \textbf{(a)} Public filings, portfolio pages, company
disclosures, and news can support reproducible analysis without subscription-only
venture data. \textbf{(b)} In the chip-company implementation, sustained product,
customer, and supply-chain progress is associated with better observed outcomes;
financing participation alone does not establish continuing operating progress.
\textbf{(c)} Continued financing accounts for 79\% of confirmed publicly visible
post-investment actions. The evidence tentatively favors considering strategic
corporate investors with relevant acquisition histories for acquisition-oriented
founders, and financing-led institutional VCs with fewer observable control events
for independence-oriented founders.

These results offer development-stage and observational guidance, not causal
guarantees or evidence of large-scale deployment performance; they neither rank
investors nor constitute investment advice. Beyond the empirical findings, we release an end-to-end
reproducibility stack---shared ontology, provenance-bearing EventChain data,
schemas, benchmarks, and executable skills---so the complete workflow can be
audited, reused, and extended.

\end{abstract}


\maketitle

\hypersetup{
  pdftitle={From Ideas to Actions: A Public-Data Decision-Support Toolchain Across the Venture Lifecycle},
  pdfauthor={Lei Qu},
  pdfsubject={Startup evaluation and retrospective investor-behavior analysis},
  pdfkeywords={startup viability, business-model prediction, venture capital,
    investor behavior, knowledge-graph event chains, public data}
}
\pdfinfo{
  /Author (Lei Qu)
  /Keywords (startup viability, business-model prediction, venture capital, investor behavior, knowledge-graph event chains, public data)
}

\section{Introduction}
\label{sec:intro}

Founders make consequential decisions under unusually weak information. Before
formation, they must decide whether a proposed business model warrants years of
effort. After funding, they must choose external operating actions toward the
next financing milestone and decide which capital partners to approach based on
their objectives and those investors' observed behavior. Commercial data
products partly address these questions, but their cost, access restrictions, and
redistribution terms limit independent scrutiny. This paper asks how much useful
decision support can be assembled from public evidence while preserving an
auditable trail.

We make three contributions. First, we describe an idea-stage evaluation pipeline
that converts a proposal into a structured candidate card, retrieves time-bounded
market and defensibility evidence, and produces a deterministic three-level verdict.
Its retrospective benchmark includes a row-disjoint independent validation sample
and out-of-sample aggregation experiments, while documenting limits to
deployment validity. Second, a 29-vendor study of the AI-inference market supplies
the five-dimensional representation used by the pipeline and exposes the disclosure
uncertainty behind each market claim. Third, we
construct versioned investor--company event records and temporally ordered chains
and use them for two exploratory
analyses: a registry of label-blind event-chain patterns and a description of dated,
announcement-visible post-investment activity.

The contribution is methodological and infrastructural rather than causal.
The work shows how public records, explicit schemas, leakage checks, and reproducible
rules can be combined. Part~Two identifies retrospective associations between
recorded actions and company outcomes, not causal effects of those actions.

Part~One (Sections~\ref{sec:part1-method}--\ref{sec:part1-casestudy})
presents the pre-founding framework, development benchmark, and market case
study. Part~Two (Sections~\ref{sec:part2-data}--\ref{sec:part2-behavior}) presents
event collection and chain construction, the exploratory pattern registry, and
investor-behavior analysis.
Sections~\ref{sec:conclusion}--\ref{sec:data} synthesize the shared conclusions,
limitations, and release information.

\section{Background and Motivation}
\label{sec:background}

\subsection{Two Information Gaps in the Startup Lifecycle}
\label{sec:bg-gaps}

At founding, a team has a product, customer, and monetization hypothesis but no
execution history. After funding, it must choose external operating actions
toward the next financing milestone and capital partners whose behavior fits
its objectives, often without transparent histories to inform either decision.
Investors face the mirror
problems of screening proposals and distinguishing capital continuity from
operational engagement. Although relevant signals are concentrated in costly
venture databases and private experience, some can be reconstructed from free
public evidence under explicit provenance and time bounds. Coverage, selection,
and retrospective labels still limit the result.

\subsection{A Methodological Choice: Free Public Data Only}
\label{sec:bg-data-principle}

The released analytical tables do not redistribute rows from paid databases.
Freely available public sources provide the reconstruction inputs at
every layer---SEC EDGAR Form~D and 13D/13G filings~\cite{sec_edgar},
Wikidata~\cite{wikidata}, the MIT-licensed OpenSporks Crunchbase free database
snapshot downloaded from Hugging Face~\cite{opensporks_crunchbase},
corporate portfolio pages, S-1 prospectuses, analyst 13F holdings, national-fund
disclosures, and a public search API for news
retrieval~\cite{tavily}. This is a
reproducibility and distribution decision. Without equivalent access, a third
party cannot independently audit, reproduce, or legally redistribute a result
built on a \$30K--\$50K-per-year database subscription. Preserved public-source
inputs make independent scrutiny and procedural reproduction more feasible,
subject to source-specific reuse terms. The release documents source
relationships, collection decisions, intended uses, and limitations in the
spirit of established dataset-documentation practice
\cite{gebru2021datasheets,pushkarna2022datacards}; its row-level source links
operationalize the provenance relation emphasized by W3C PROV
\cite{w3c_prov_dm}.

The choice imposes measurable coverage costs. In Part~Two, reliance on
announcement-level records means interface events between
investors and companies are only $\sim$70\% covered by free sources, per-investor
amounts are populated for only $\sim$16\% of rows, and sovereign-wealth coverage
is $\sim$50\% (Section~\ref{sec:general-limitations}). Although the OpenSporks
snapshot is distributed under MIT, its records originate from Crunchbase; we
therefore do not redistribute the Part~One row-level mirror data and publish only
aggregate benchmark statistics and methodology (Section~\ref{sec:data}).
These observation limits bound the inferences the reconstruction can support.

\subsection{Why the Two Parts Form One Toolchain}
\label{sec:bg-why-together}

\textbf{Two stages of founder decision assistance.} Part~One asks whether a
proposed business model warrants pursuit before founding. Part~Two examines
which operating-action patterns and capital-partner behaviors are observed
along different post-founding company trajectories. It provides retrospective
evidence for those decisions, not a forecast of a particular investor's conduct.
The workflows serve the same founder at different stages; a Part~One verdict
does not feed Part~Two.

\textbf{Shared evidence discipline.} Both parts use time bounds, provenance, and
task-specific leakage controls. Part~One restricts public-market evidence to the
three years preceding the founding year through the founding year
(Section~\ref{sec:p1m-retrieval}) and removes outcome leakage from inputs before
evaluation (Section~\ref{sec:p1b-leakage}). For confirmed post-investment
statistics, Part~Two requires interface events to occur strictly after investment
(Section~\ref{sec:p2-behavior-method}), while its
pattern grammar generates hypotheses without outcome labels
(Section~\ref{sec:p2-pattern-method}). Part~Two analyzes completed chains and
explicitly permits terminal-node attributes where a registered pattern calls for
them; it should not be read as a point-in-time forecasting design.

\textbf{Shared infrastructure.} Both parts turn heterogeneous public records
into versioned, auditable artifacts. The shared outcome ontology supplies
benchmark truth in Part~One and company labels for retrospective association in
Part~Two; it does not label operating actions or investors. The skill repository
packages both workflows, while the released EventChain dataset supports the
Part~Two analysis.

\begin{figure}[!t]
\centering
\includegraphics[width=0.94\textwidth]{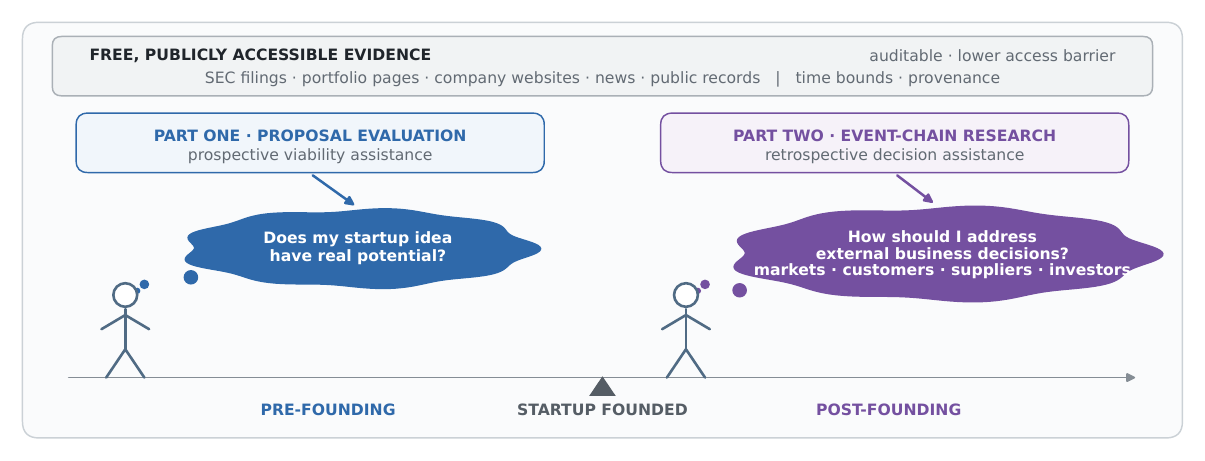}
\Description{A founder moves along a timeline from asking whether a startup idea has potential before founding to asking how to handle external market, customer, supplier, and investor decisions after founding. Part One and Part Two provide assistance at the respective stages, supported by freely available public evidence.}
\caption{One founder-decision-assistance envelope across the startup lifecycle.
Part~One addresses the pre-founding proposal decision; Part~Two supplies retrospective
evidence for post-founding external business decisions involving markets, customers,
suppliers, and investors. Both use freely available public
sources rather than requiring subscription-only venture data.}
\label{fig:decision-lifecycle}
\end{figure}
\FloatBarrier

Figure~\ref{fig:decision-lifecycle} locates the two decision stages around company
formation.

\subsection{Who This Work Is For}
\label{sec:bg-audience}

Founding teams receive pre-pitch viability evidence and retrospective evidence
for external operating actions and capital-partner selection. Investors and analysts receive a precision-weighted
screening benchmark and auditable event-chain hypotheses. Data engineers receive
versioned skills and EventChain data demonstrating provenance-bearing entity
resolution, cross-source fusion, and label-blind analysis.

\subsection{Related Work and Positioning}
\label{sec:bg-related}

\textbf{Startup business-model prediction.} Prior data-driven systems predict
startup outcomes from structured venture-platform records~\cite{corea2021}, while
a recent language-model system combines Crunchbase fundamentals with textual
self-descriptions~\cite{maarouf2025}. Longitudinal evidence also shows that firms'
business lines are more stable from early plans to public-company status than
their management teams, clarifying what an idea-stage representation can and
cannot preserve~\cite{kaplan_sensoy_stromberg2009}. VCBench is a conceptually
related and active community for predicting startup outcomes from \emph{founder
profiles}. At the time of access, GPT-4o has the highest $F_{0.5}$ among the
general-purpose model baselines (0.257), while Verifiable-RL, which learns a
task-specific policy, leads the overall leaderboard at 0.366
\cite{vcbench,think_reason_learn}. VCBench differs from our study in data type, population
scope, and prediction target. These results are therefore not directly comparable
with ours, and our higher $F_{0.5}$ should not be interpreted as evidence that
this pipeline outperforms systems on the VCBench leaderboard. We cite VCBench
because the two prediction problems are conceptually related and VCBench is one
of the most active communities studying this broader class of problems
(Section~\ref{sec:p1b-baselines}). PHBench finds that
\emph{launch-day} signals carry statistically significant predictive
information for Series~A within 18 months~\cite{phbench}, supporting our
product-launch $\rightarrow$ outcome patterns
(Section~\ref{sec:p2p-p1p8}). Meta-analytic evidence ranks traction and firm
characteristics as the strongest predictor families across thirteen
studies~\cite{startup_success_meta_analysis}. There is also a substantial
method-adjacent literature on event-structured prediction---neural point
processes over fundraising events~\cite{cta_npp_2025} and graph-augmented
time-series prediction---as well as large industrial knowledge graphs of
companies and investors~\cite{companykg}. Part~One differs in three respects.
First, it restricts input to the \emph{idea-stage proposal} and time-bounded
public-market context, matching the pre-pitch decision. Second, it applies a
leakage-control layer and audits its precision and under-detection
(Section~\ref{sec:p1b-leakage}). Third, it distinguishes the 198-row
threshold-tuning development sample from the independently drawn, row-disjoint
198-row validation sample within the combined 396-row benchmark, and separately draws a 4{,}000-row
scale-validation cohort.
Among the completed cases, a 377-row five-sublabel composition-matched subset
yields $F_{0.5}=0.6573$ [0.556, 0.746], while the 1{,}199-row reserved split
remains untouched (Section~\ref{sec:general-limitations}).

\textbf{Investor behavior.} A broad equity-financing literature distinguishes
screening, contracting, monitoring, value-add, and exit across investor
types~\cite{drover2017}. Survey evidence likewise shows that VC decisions span
both investment selection and post-investment activity~\cite{gompers2020}.
This literature provides interpretive anchors for Part~Two rather than its method.
Evidence that closer VC monitoring changes portfolio-company outcomes
underscores why dated post-investment activity matters~\cite{bernstein2016}.
The finding that founder-CEO replacement has a negative naive correlation with
performance but a positive instrumental-variable estimate~\cite{ewens_marx_2018}
motivates a context-dependent interpretation of our ownership-change result:
major-holder changes are negatively associated with outcomes only in the observed
low-follow-through contexts, rather than being uniformly adverse
(Section~\ref{sec:p2p-p1p8}). The finding that corporate venture
capital creates firm value only under strategic, not purely financial,
motives~\cite{dushnitsky_lenox_2006} aligns with our observation that CVCs
depart from a financing-only profile through exits, strategic investments,
and board changes (Section~\ref{sec:p2b-types}). The syndication literature,
which documents an inverted-U relationship between single-round syndicate size and
performance~\cite{vc_syndicate_size_2021}, provides the relevant comparison for
our P2 result, which counts investors at the terminal chain-boundary event
when no intermediate action is recorded
(Section~\ref{sec:p2p-p1p8}). Where
practitioner surveys emphasize how much hands-on value VC firms claim to add,
our evidence reads the \emph{observable, dated record}, and we flag the
resulting absence-of-evidence asymmetry as a limitation rather than a finding.

\textbf{Public-data investor graphs.} Reconstructing investor$\rightarrow$company
graphs from public filings has established precedents, including EDGAR-derived
ownership graphs, open startup graphs combining SEC data with other free
sources, and the peer-reviewed industrial CompanyKG~\cite{companykg}. Part~Two
differs in granularity and purpose. It fuses
\emph{funding events} (Form~D, SPV timing, news, S-1, 13D/13G) with
\emph{portfolio pages} into a single dated event graph with provenance
(Section~\ref{sec:p2d-edges}), then subjects the graph to
\emph{label-blind pattern enumeration} to produce a catalog of retrospective
complete-chain associations (23{,}308 patterns, with an interpretable shortlist
reported in Section~\ref{sec:p2p-p1p8}), plus an investor-side behavior
comparison using confirmed post-investment rows and support-gated investor histories. Temporal-graph
benchmarks likewise emphasize dated nodes or edges and explicit evaluation
protocols~\cite{huang2023tgb}; EventChain contributes a provenance-bearing
descriptive event record and pattern workflow rather than a temporal-graph
prediction leaderboard.

\textbf{Scope of the contribution.} We do not claim to be
the first to predict startup outcomes, to describe investor behavior, or to
build a public-data investor graph; each has an established literature. The
contributions described in Section~\ref{sec:intro} combine a validated
proposal-evaluation pipeline, an industry study of its representation and use,
and an auditable event-and-chain workflow for retrospective analysis, all built
from freely available public evidence.
Sections~\ref{sec:p1b-limits} and
\ref{sec:p2b-limits} define the corresponding limits.


\section{Shared Outcome Ontology and Label Semantics}
\label{sec:shared-outcomes}

\begin{sloppypar}
Both parts use the same deterministic company-outcome ontology. It is computed
from structured company records independently of the Part~One evaluation framework
and the Part~Two pattern engine. The fine-grained field
\texttt{outcome\_label\_v4} maps to the three-level company field
\texttt{label\_v4} as follows.
\end{sloppypar}

Table~\ref{tab:outcome-ontology} gives the complete mapping.

\begin{center}
\begin{minipage}{\textwidth}
\centering
\captionof{table}{Shared company-outcome ontology used in Parts One and Two.}
\label{tab:outcome-ontology}
\small
\renewcommand{\arraystretch}{1.12}
\setlength{\tabcolsep}{4pt}
\begin{tabular}{>{\raggedright\arraybackslash}p{0.39\linewidth}>{\raggedright\arraybackslash}p{0.18\linewidth}>{\raggedright\arraybackslash}p{0.34\linewidth}}
\toprule
Fine-grained outcome & Company label & Operational criterion \\
\midrule
\texttt{POSITIVE\_IPO} & SUCCESS & Public listing \\
\texttt{POSITIVE\_LATE\_STAGE\_FUNDED} & SUCCESS & Active and reached Series B or later, private equity, or post-IPO equity \\
\texttt{POSITIVE\_ACQUIRED} & SUCCESS & Acquisition identified in the structured-record description \\
\texttt{EQUIVOCAL\_DELISTED} & SUCCESS & Delisted; grouped with VC-style exits by default and retained separately for sensitivity analysis \\
\texttt{NEGATIVE\_CLOSED} & FAILURE & Explicitly closed \\
\texttt{NEGATIVE\_NO\_TRACTION} & FAILURE & Active but without Series-A-level funding at least five years after founding and last funding \\
\texttt{INDETERMINATE} & AMBIGUOUS & Active but no committed positive or negative milestone \\
\texttt{UNKNOWN} & AMBIGUOUS & Required operating and IPO status fields unavailable \\
\bottomrule
\end{tabular}
\end{minipage}
\end{center}

SUCCESS and FAILURE are financing/exit milestone proxies, not measurements of
profitability, product quality, social value, or long-run survival. AMBIGUOUS
companies are retained for coverage reporting but excluded whenever a binary
SUCCESS--FAILURE comparison is made.

Three label namespaces must remain distinct. \texttt{label\_v4} classifies a
\emph{company outcome}. Part~One's \texttt{PASS}/\texttt{WARN}/\texttt{FAIL}
labels are framework verdicts that are evaluated against that outcome proxy.
Part~Two's \texttt{RECOMMEND}/\allowbreak\texttt{AVOID}/\allowbreak\texttt{INDIFFERENT}/\allowbreak%
\texttt{INCONCLUSIVE} labels classify an observed operating action or event-chain
pattern by the direction and threshold status of its retrospective association
with company outcomes. They never classify a company. In particular,
\texttt{RECOMMEND} and \texttt{AVOID} are preserved registry terms, not causal
treatment recommendations, prospective forecasts, or investment advice.

\par\noindent\begin{minipage}{\textwidth}
\researchpart{One}{Pre-Founding Decision Assistance}

Before founding, an aspiring entrepreneur faces a central decision:
\textbf{Does this business idea warrant pursuit?} Part~One evaluates proposals
through public-evidence market and moat checks, followed by deterministic
aggregation (Figure~\ref{fig:part1-architecture}).

\begin{center}
\centering
\includegraphics[width=\textwidth]{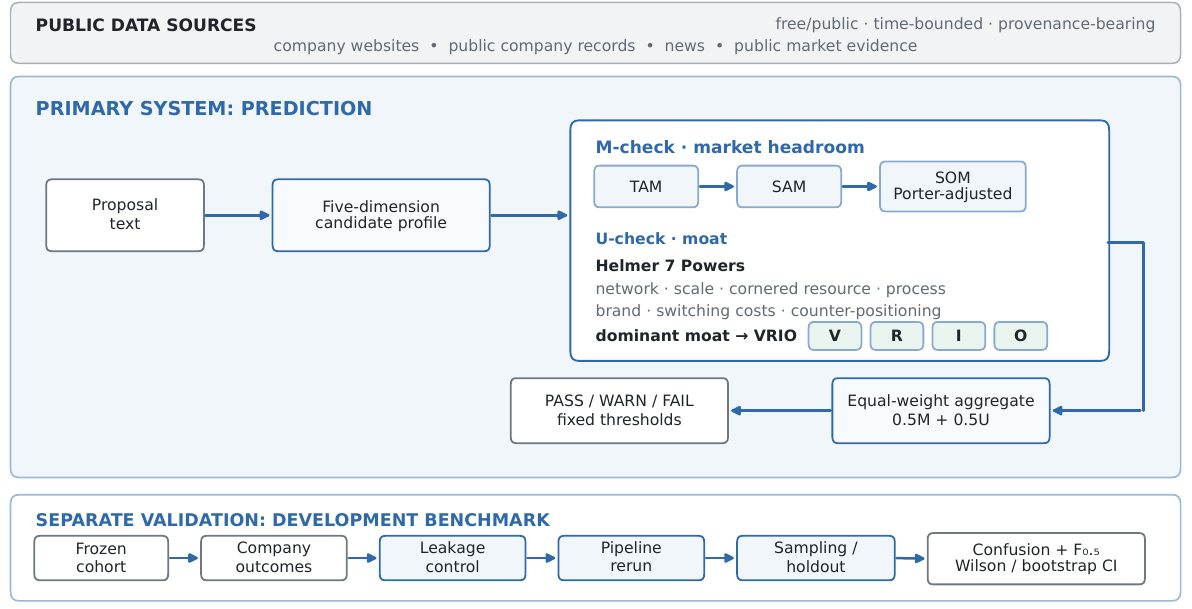}
\Description{Part One architecture: proposal profiling, market and moat checks, deterministic aggregation, verdict, and separate development benchmark.}
\ifdefined\captionsetup\captionsetup{hypcap=false}\fi
\captionof{figure}{Part~One architecture: prediction pipeline and benchmark validation.}
\label{fig:part1-architecture}
\end{center}
\end{minipage}\par

Section~\ref{sec:part1-method} presents the prediction pipeline,
Section~\ref{sec:part1-benchmark} evaluates it against historical company outcomes,
and Section~\ref{sec:part1-casestudy} examines the AI-inference vendor market as
an industry case study.
\section{Method: Proposal Evaluation Pipeline}
\label{sec:part1-method}

The pipeline treats a proposal's claims as hypotheses to be checked, not as
predetermined conclusions. Market and moat checks assess these claims against
public evidence under explicit evaluation criteria. A deterministic aggregation
rule combines the check outputs into a \texttt{PASS}, \texttt{WARN}, or
\texttt{FAIL} verdict.

\subsection{Task and Input Signal}
\label{sec:p1m-task}

We study the problem of \emph{business-model prediction at the idea stage}: given
the free-text description of a proposed startup (its product category, target
customers, and monetization hypothesis), predict whether the venture will reach a
positive outcome. The intended prediction is made \emph{as if at founding}. In
the retrospective benchmark, however, candidate cards can include present-day
category tags, taglines, and website text. Leakage controls remove explicit
outcome cues, but cannot prove that these inputs reproduce the information
distribution of a genuine founding-time proposal. The benchmark therefore tests
a development proxy for the intended pre-pitch use, not that deployment setting.

The benchmark target is the shared operational funding/exit proxy defined in
Section~\ref{sec:shared-outcomes}. Part~One predicts that company-level proxy; its
\texttt{PASS}/\texttt{WARN}/\texttt{FAIL} verdicts are framework outputs rather
than outcome labels. AMBIGUOUS outcomes are excluded from binary scoring.

\subsection{Candidate-Card Representation}
\label{sec:p1m-card}

Raw proposals are heterogeneous. We therefore normalize each proposal into a
structured \emph{candidate card}---a five-section distillation produced by the
candidate-profiler component:

\begin{itemize}
  \item \textbf{Revenue model} --- primary and secondary revenue streams, pricing
        structure, and a plausibility assessment;
  \item \textbf{Customer segmentation} --- identifiable segments, estimated mix, and
        disclosed logos (if any);
  \item \textbf{Cost structure} --- a best-effort estimate of the business's main
        production or service-delivery costs, R\&D, sales and marketing, and gross margin;
  \item \textbf{Differentiation and moat} --- claimed advantages assessed using
        Helmer's Seven Powers taxonomy~\cite{helmer_7powers};
  \item \textbf{Strategic vulnerabilities} --- the principal risks to the model.
\end{itemize}

Appendix~\ref{app:candidate-card} illustrates the five dimensions with a fictional
proposal, distinguishing stated assumptions from missing information.

The card is assembled from the proposal text, the company's category tags, a
tagline, and (where available) website body text. Crucially, the card is written in
\emph{idea-stage narrative}: it summarizes the business model \emph{as proposed},
not the company's realized trajectory. A separate leakage-scan step
(Section~\ref{sec:p1b-leakage}) removes residual outcome hints (retrospective
language, ticker symbols, named anchor-customer logos, dated claims), reducing
the risk that the framework can read the answer directly from the input.

\subsection{The \texttt{evaluate-proposal} Framework}
\label{sec:p1m-framework}

The evaluation is an orchestrated multi-step pipeline, implemented as a reusable
agent skill (\texttt{evaluate-proposal}) that drives three sibling components in a
fixed order. It is methodologically adjacent to retrieval-augmented generation
\cite{lewis2020rag} and interleaved reasoning-and-action agents
\cite{yao2023react}, but replaces an open-ended agent trajectory with typed
artifacts and deterministic aggregation:

\begin{enumerate}
  \item \textbf{Candidate profiling} --- the candidate profiler
        (Section~\ref{sec:p1m-card}) converts the proposal into a structured
        five-dimensional card. It does not assign an archetype; the seven-category
        vendor taxonomy in Section~\ref{sec:p1cs-arch} is retained only as
        descriptive case-study context.
  \item \textbf{M-check (market headroom)} --- estimates total addressable market
        (TAM), serviceable addressable market (SAM), and serviceable obtainable
        market (SOM) headroom via time-bounded public retrieval, then applies a
        Porter five-forces discount (rivalry, substitutes, entry, buyer and supplier
        power) and an incumbent-lock factor to arrive at a \emph{headroom band}.
        The check outputs a verdict and a structured evidence record
        (\texttt{m\_check.json}).
  \item \textbf{U-check (moat)} --- evaluates the candidate's
        claimed moats with a VRIO analysis (Value, Rarity, Inimitability,
        Organization), again grounded in time-bounded competitive-landscape
        retrieval, and outputs a verdict and structured record
        (\texttt{u\_check.json}).
  \item \textbf{Aggregation} --- a deterministic rule maps the two check verdicts to a
        final verdict. Each verdict maps to a score
        ($\mathrm{PASS}=1.0$, $\mathrm{WARN}=0.5$, $\mathrm{FAIL}=0.0$) and
        $\mathrm{score}=0.5\,M+0.5\,U$; the thresholds are $\ge 0.5 \Rightarrow
        \mathrm{PASS}$, $\ge 0.25 \Rightarrow \mathrm{WARN}$, and otherwise
        $\mathrm{FAIL}$. The rule is deliberately simple --- it is a fixed,
        versioned composition whose sole free parameter is the PASS threshold.
  \item \textbf{Reasoning composition} --- a narrative layer synthesizes the two
        evidence records and the aggregate verdict into a human-readable
        recommendation with a dominant-driver attribution.
\end{enumerate}

\textbf{Defense against input manipulation.} The pipeline is designed to
evaluate the proposal against the market and moat rules, not to adopt its
self-reported assessments. M-check requires external triangulation of market
size; U-check tests claimed moats against VRIO criteria and competitive evidence.
The prompts are designed to maintain this distinction and reduce the risk that
an input can manipulate the pipeline by embedding purported check results or
desired verdicts. Such statements are claims to examine, not completed checks.

The reported Part~One evaluation configuration is dated June 2026. It uses a
leading general-purpose LLM with extended thinking/reasoning disabled. The
released workflow fixes the prompts, schemas, retrieval constraints, component
order, and deterministic aggregation rule; the model-independent artifact
contracts, rather than a provider-specific interface, define the pipeline.

The three verdict levels carry distinct meanings. \textbf{PASS} means
``recommend'' (the framework expects success), \textbf{WARN} indicates
abstention, and \textbf{FAIL} means ``do not recommend.'' Binary evaluation treats
only \texttt{PASS} as positive; \texttt{WARN} and \texttt{FAIL} are rejections.

\begin{figure*}[t]
\centering
\begin{adjustbox}{max width=0.74\textwidth,center}
\begin{tikzpicture}[x=1.48cm,y=0.92cm,font=\figurebodyfont]
  \node[rotate=90,font=\figuretitlefont] at (-1.15,1.5) {M-check verdict};
  \node[font=\figuretitlefont] at (1.5,4.05) {U-check verdict};
  \foreach \i/\lab in {0/FAIL,1/WARN,2/PASS} {
    \node at (-0.42,\i+0.5) {\lab};
    \node at (\i+0.5,3.48) {\lab};
  }
  \foreach \x/\y/\fill/\verdict/\score in {
    0/0/reportRed!22/FAIL/0.00,
    1/0/reportOrange!27/WARN/0.25,
    2/0/reportGreen!24/PASS/0.50,
    0/1/reportOrange!27/WARN/0.25,
    1/1/reportGreen!24/PASS/0.50,
    2/1/reportGreen!34/PASS/0.75,
    0/2/reportGreen!24/PASS/0.50,
    1/2/reportGreen!34/PASS/0.75,
    2/2/reportGreen!44/PASS/1.00} {
    \filldraw[fill=\fill,draw=white,line width=1.2pt] (\x,\y) rectangle +(1,1);
    \node[font=\figuretitlefont] at (\x+0.5,\y+0.62) {\verdict};
    \node[font=\figurenotefont,text=black!65] at (\x+0.5,\y+0.32) {score \score};
  }
  \node[anchor=west,align=left,font=\figurebodyfont] at (3.35,2.52)
    {$\mathrm{score}=0.5M+0.5U$};
  \node[anchor=west,align=left,font=\figurebodyfont] at (3.35,1.65)
    {$\geq0.50$: PASS\\$\geq0.25$: WARN\\otherwise: FAIL};
  \node[anchor=west,align=left,font=\figurenotefont,text=black!70] at (3.35,0.40)
    {Only FAIL + FAIL\\produces a final FAIL.};
\end{tikzpicture}
\end{adjustbox}
\Description{A three-by-three matrix maps every M-check and U-check verdict pair to its numeric aggregate score and final PASS, WARN, or FAIL verdict.}
\caption{Complete decision surface of the released equal-weight aggregator.
Rows are M-check verdicts and columns are U-check verdicts. The cell reports the
aggregate score and resulting framework verdict; this is a fixed rule, not a
learned probability surface.}
\label{fig:decision-surface}
\end{figure*}
\FloatBarrier

Figure~\ref{fig:decision-surface} exposes the complete deterministic decision
surface rather than leaving the thresholds as prose. The current rule is permissive
once either check passes: only two FAIL checks produce a final FAIL, while a
PASS paired with a FAIL still reaches the 0.50 PASS threshold. This behavior is
subsequently tested, rather than assumed effective, in Section~\ref{sec:p1b-baselines}.

\subsection{Time-Bounded Retrieval and Leakage Control}
\label{sec:p1m-retrieval}

Both checks ground their evidence in public-market retrieval using a web search API,
\emph{restricted to a window of three years before founding through founding year}.
The queries are constructed for \emph{market context} --- category size, competitor
landscape, technology trajectory --- not for the company's own post-founding
outcome. This time-bounding improves reproducibility and blocks a direct class of
post-founding evidence. It does not by itself establish that retrospective cards
match genuine proposal inputs, so benchmark performance should not be assumed to
transfer unchanged to deployment. Leakage control operates at three layers:
the idea-stage distillation of the card, the regex and structural leakage sweep, and the
retrieval time-lock.

The retrieval lock constrains query terms and accepted evidence dates, but it is
not a historical-web archive. A page retrieved today can describe an earlier
market period using information published later. Without archived page snapshots
and publication-date verification for every item, residual point-in-time leakage
cannot be excluded.

\subsection{Released Operating Modes}
\label{sec:p1m-prompts}

The released skill repository separates orchestration, profiling, and benchmark
validation:

\begin{itemize}
  \item \texttt{evaluate-proposal} --- the full single-candidate pipeline
        (card $\rightarrow$ M-check $\rightarrow$ U-check $\rightarrow$
        aggregate $\rightarrow$ reasoning);
  \item \texttt{candidate-profiler} --- card construction and validation when
        only the proposal representation is required;
  \item \texttt{benchmark-validation} and \texttt{leakage-scan} --- cohort,
        holdout, metric, and temporal-integrity tools for retrospective testing.
\end{itemize}

The benchmark utilities evaluate the same proposal-level verdict contract, but
they are separate from the runtime predictor and do not add a company or
archetype classification stage.

\section{Engineering Validation: Pipeline Benchmark}
\label{sec:part1-benchmark}

We evaluate the framework's association with outcomes using development,
row-disjoint validation, and independently drawn scale-cohort evidence built
from freely accessible data. This section reports the benchmark
construction, the observed iteration plateau, LLM and conventional baselines, system-level
ablation, and the limitations of the current evidence.

\subsection{Cohort and Metric}
\label{sec:p1b-cohort}

\textbf{Cohort.} We use the MIT-licensed OpenSporks Crunchbase free database
snapshot downloaded from Hugging Face~\cite{opensporks_crunchbase}
(2.87M raw company records) and filter to an \emph{AI-adjacent} cohort: companies
whose structured category tags match AI, ML, robotics, analytics, computer vision,
natural language processing, big data, or generative AI, founded 2010--2024. The resulting cohort has
\emph{37{,}569 rows}. This is an AI-adjacent technology-startup population, not a
general-startup population; 87.8\% of rows explicitly carry an ``Artificial
Intelligence (AI)'' tag.

Applying the shared outcome ontology of Section~\ref{sec:shared-outcomes}, SUCCESS
is 4.09\%,
FAILURE 42.92\%, and AMBIGUOUS 53.00\%. We additionally study a \emph{chip-only}
subset (233 rows, category-tagged semiconductor) and a hardware subset (5{,}298
rows); chip startups show a materially higher SUCCESS rate (17.17\%), consistent
with the capital intensity and long development cycles of the category.

The reported 396-row benchmark is not a prevalence-matched draw from this cohort.
It combines a 198-row development sample (seed 43) and a separately drawn,
row-disjoint 198-row validation sample (seed 44), both outcome-stratified. Each
contains 48 SUCCESS cases (42 late-stage funded, three
IPO, and three acquired) and 150 FAILURE cases (129 no-traction and 21 closed),
for 96 SUCCESS and 300 FAILURE cases overall. Sampling required an already frozen
candidate card and excluded AMBIGUOUS outcomes. The resulting 24.2\% SUCCESS share
was chosen to provide enough positive cases for development comparison; because
precision and $F_{0.5}$ depend on class prevalence, the reported scores describe
this stratified pool and are not deployment-prevalence estimates for the 4.09\%
SUCCESS cohort.

\begin{figure*}[t]
\centering
\includegraphics[width=0.90\textwidth]{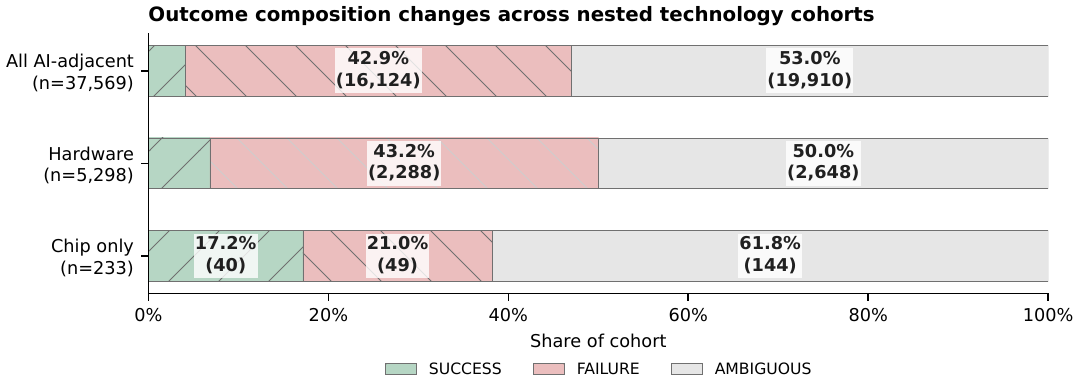}
\Description{Stacked bars showing SUCCESS, FAILURE, and AMBIGUOUS shares and counts in the full AI-adjacent cohort, hardware subset, and chip-only subset.}
\caption{Outcome composition in the three nested benchmark cohorts. Counts are
shown inside segments where space permits. The chip-only subset has a higher
SUCCESS share and a larger AMBIGUOUS share than the full AI-adjacent cohort;
these composition differences motivate reporting the population explicitly.}
\label{fig:cohort-outcomes}
\end{figure*}
\FloatBarrier

Figure~\ref{fig:cohort-outcomes} shows that the chip subset is not merely smaller: its
observed outcome mix differs sharply from the broader cohort, while AMBIGUOUS
remains the largest class.

\textbf{Metric.} We score the framework with the $F_{0.5}$ measure,
\[
F_\beta=\frac{(1+\beta^2)\,P\,R}{\beta^2\,P+R},
\]
with $\beta=0.5$, weighting precision more heavily than recall. This encodes a
deliberately conservative screening policy: acting on a false positive can commit
capital and founder time, while a false negative is also consequential but is
treated by this operating mode as a request for further evidence rather than an
automatic abandonment decision. The weighting is a product-policy choice, not a
universal founder utility function. Evaluations are binary: only \texttt{PASS} is a positive
prediction; \texttt{WARN} and \texttt{FAIL} count as rejections
(Section~\ref{sec:p1m-framework}). AMBIGUOUS-truth rows are filtered before scoring.

\subsection{Data Quality and Leakage Control}
\label{sec:p1b-leakage}

All $37{,}569$ cohort rows were preprocessed into candidate cards
(Section~\ref{sec:p1m-card}), frozen as a permanent data asset. A data-quality phase
then scanned every card for \emph{outcome-label leakage}: content that reveals the
company's realized fate (retrospective language, ticker symbols, named
anchor-customer logos, dated claims). Leakage was detected and removed while keeping
the row; affected embeddings were regenerated. The scan found high-severity leakage
in 4.3\% of the benchmark sample; the SUCCESS-group rate was 1.8$\times$ the
FAILURE-group rate, confirming that leakage removal was necessary to obtain
less contaminated performance estimates. A companion audit estimated over-detection
at 0.36\% and under-detection at 5.7\%.

\subsection{Iteration History and the Observed \texorpdfstring{$F_{0.5}$}{F0.5} Plateau}
\label{sec:p1b-iterations}

The framework was developed through a recorded iteration series, each iteration
pairing a framework version with a fixed sample and reporting $F_{0.5}$ with its
recorded uncertainty range. Earlier iterations use approximate ranges propagated
from Wilson bounds on precision and recall; the clean iter-13/14 and pooled
headlines use bootstrap intervals. Once recorded, a point estimate is
never amended, so the series reads
as a calibration narrative. The main points:

\begin{itemize}
  \item \textbf{The initial baseline is zero.} The first threshold configuration
        (\texttt{PASS}$\geq0.75$) never fires on samples drawn from the full cohort: the
        observed aggregate M/U scores take values in $\{0, 0.25, 0.5\}$, so a
        threshold of 0.75 produces no PASS verdicts at all ($F_{0.5}=0$).
        Lowering the PASS threshold to 0.50 (version v1.5a) was a functional correction,
        not a bias adjustment.
  \item \textbf{The repaired framework validates on a row-disjoint sample and
        then plateaus.} With the threshold fixed, the 198-row development and
        independently drawn 198-row validation samples yielded $F_{0.5}=0.708$
        and $0.616$ on the original cards; after leakage removal, they yield
        \textbf{0.729 and 0.536}, respectively, with a pooled score of \textbf{0.6301}
        [bootstrap CI 0.537, 0.710]. Leakage removal itself changes the headline
        little (pooled leaky 0.661 vs.\ clean 0.630, with overlapping intervals); the
        time-bounded retrieval that drives the checks never consumed the leaked
        content.
  \item \textbf{Aggregation variants do not surpass the observed plateau.}
        Discrete aggregation and offline continuous rescoring yield
        $F_{0.5}\approx0.63$--$0.66$ on this pool. Lossless learned-encoding
        variants, including a 30-parameter search, overfit the 198-row training
        pools: in-sample scores reach 0.75--0.77, but 4-fold out-of-sample scores
        are 0.53--0.58, below the discrete and continuous-rescoring results.
        We therefore require out-of-sample validation before reporting
        $F_{0.5}$ for any variant with more than one hyperparameter.
\end{itemize}

Validation views are summarized in Table~\ref{tab:validation-views}. The
row-disjoint validation estimate is the clean confirmatory result for the frozen
threshold; the pooled estimate is retained as a higher-precision combined
benchmark description rather than presented as an untouched test result.

\begin{center}
\begin{minipage}{\textwidth}
\centering
\captionof{table}{Validation views for the Part One pipeline.}
\label{tab:validation-views}
\small
\renewcommand{\arraystretch}{1.12}
\begin{adjustbox}{max width=\textwidth,center}
\begin{tabular}{lccc}
\toprule
Validation view & Rows & $F_{0.5}$ & Interpretation \\
\midrule
Clean development draw (iter~13) & 198 & 0.7292 & threshold-tuning sample \\
Row-disjoint validation draw (iter~14) & 198 & 0.5357 & row-disjoint validation \\
Pooled clean benchmark & 396 & 0.6301 [0.537, 0.710] & combined benchmark estimate \\
Completed scale cases, post-stratified & 1{,}027 & 0.6506 [0.598, 0.707] & stratum-weighted sensitivity estimate \\
Independent scale-cohort subset & 377 & 0.6573 [0.556, 0.746] & composition-matched holdout check \\
Learned encodings, 4-fold out-of-sample & 396 & 0.53--0.58 & overfit diagnostic \\
\bottomrule
\end{tabular}
\end{adjustbox}
\end{minipage}
\end{center}

Table~\ref{tab:validation-views} separates the row-disjoint development-loop validation from the pooled
headline, the independently drawn scale-cohort check, and the four-fold
out-of-sample model-selection diagnostic. Post-stratification uses all 1{,}027
completed scale-cohort cases and weights each of the five outcome strata to its
fixed count in the 2{,}801-row execution split. It yields precision 0.654, recall
0.638, and $F_{0.5}=0.6506$ [stratified-bootstrap 0.598, 0.707]. This estimate
assumes that completion is exchangeable within each outcome stratum. The
deterministic 377-row composition-matched subset avoids unequal weights and
provides a conservative sensitivity check; its point estimate differs from the
pooled benchmark by only $+0.0272$, and the intervals overlap.

\begin{figure*}[t]
\centering
\begin{adjustbox}{max width=\textwidth,center}
\begin{tikzpicture}[
  x=0.70cm, y=6.0cm,
  font=\figurebodyfont,
  >=stealth,
  every node/.style={font=\figurebodyfont},
  ebar/.style={|-|, thick, reportBlue!80},
  pt/.style={fill=reportBlue, draw=reportBlue, inner sep=1.2pt, circle},
  ptmaj/.style={fill=reportBlue, draw=reportBlue, inner sep=1.2pt, circle},
]
\fill[gray!18] (0.4,0.63) rectangle (15.5,0.66);
\node[anchor=south west, gray!60] at (15.5,0.66) {observed 0.63--0.66};

\draw[dotted, reportRed!55] (0.4,0.273) -- (15.5,0.273);
\node[anchor=south west, reportRed!55] at (15.5,0.273) {Raw LLM: 0.273};

\draw[->, thick] (0.4,0) -- (15.6,0) node[below, midway, yshift=-1.2em] {Iteration};
\draw[->, thick] (0.4,0) -- (0.4,1.02);
\node[anchor=east, xshift=-6pt] at (-0.22,0.50) {$F_{0.5}$};

\foreach \y/\lab in {0.0/0.0,0.1/0.1,0.2/0.2,0.3/0.3,0.4/0.4,0.5/0.5,0.6/0.6,0.7/0.7,0.8/0.8,0.9/0.9,1.0/1.0}{
  \draw[black!25, very thin] (0.4,\y)--(15.5,\y);
  \node[left] at (0.38,\y) {\lab};
}
\draw[thick, black!70] (0.4,0)--(0.4,1.0);

\foreach \x/\lab in {1/1,2/2,3/3,4/4,5/5,6/6,7/7,8/8,9/9,10/10,11/11,12/12,13/13,14/14}{
  \draw[black!55, very thin] (\x,0)--(\x,-0.015);
  \node[below] at (\x,-0.02) {\lab};
}

\draw[thick, reportBlue] plot[smooth] coordinates {
  (1,0.000) (2,0.250) (3,0.000) (4,0.708) (5,0.000) (6,0.616)
  (7,0.655) (8,0.637) (9,0.689) (10,0.650) (11,0.542)
  (12,0.578) (13,0.729) (14,0.536)};

\node[pt] at (2,0.250) {};
\draw[ebar] (4,0.568)--(4,0.818); \node[pt] at (4,0.708) {};
\draw[ebar] (6,0.482)--(6,0.734); \node[pt] at (6,0.616) {};
\draw[ebar] (7,0.607)--(7,0.700); \node[pt] at (7,0.655) {};
\draw[ebar] (8,0.567)--(8,0.701); \node[pt] at (8,0.637) {};
\draw[ebar] (9,0.621)--(9,0.749); \node[pt] at (9,0.689) {};
\draw[ebar] (10,0.601)--(10,0.695); \node[pt] at (10,0.650) {};
\draw[ebar] (10.8,0.481)--(10.8,0.579); \node[pt] at (10.8,0.530) {};
\draw[ebar] (11.2,0.504)--(11.2,0.602); \node[pt] at (11.2,0.554) {};
\draw[ebar] (11.8,0.525)--(11.8,0.622); \node[pt] at (11.8,0.574) {};
\draw[ebar] (12.2,0.532)--(12.2,0.629); \node[pt] at (12.2,0.581) {};
\draw[ebar] (13,0.606)--(13,0.837); \node[pt] at (13,0.729) {};
\draw[ebar] (14,0.412)--(14,0.655); \node[pt] at (14,0.536) {};

\node[pt] at (1,0.0) {};
\node[pt] at (3,0.0) {};
\node[pt] at (5,0.0) {};
\node[anchor=north west, reportBlue] at (15.5,0.6301) {Full Pipeline: 0.6301};

\draw[dashed, reportBlue!75] (0.4,0.6301) -- (15.5,0.6301);
\end{tikzpicture}
\end{adjustbox}
\Description{Line plot of F0.5 across fourteen framework iterations, including approximate uncertainty intervals and the observed development-pool plateau.}
\caption{Iteration history: $F_{0.5}$ per iteration with recorded uncertainty
ranges (approximate precision/recall-propagated ranges for earlier iterations and
bootstrap intervals for clean iter~13/14). Each iteration uses its recorded
sample; iterations 13 and 14 use separate 198-row development and validation
samples. The full-pipeline reference of 0.6301 combines those two samples
(396 rows). The figure also shows the zero baselines at iterations 1, 3, and 5,
the observed development-pool band $0.63$--$0.66$, and the internal Raw LLM
baseline. Iteration 11/12 variants (a/b) are plotted at their
recorded micro-positions.}
\label{fig:iterations}
\end{figure*}
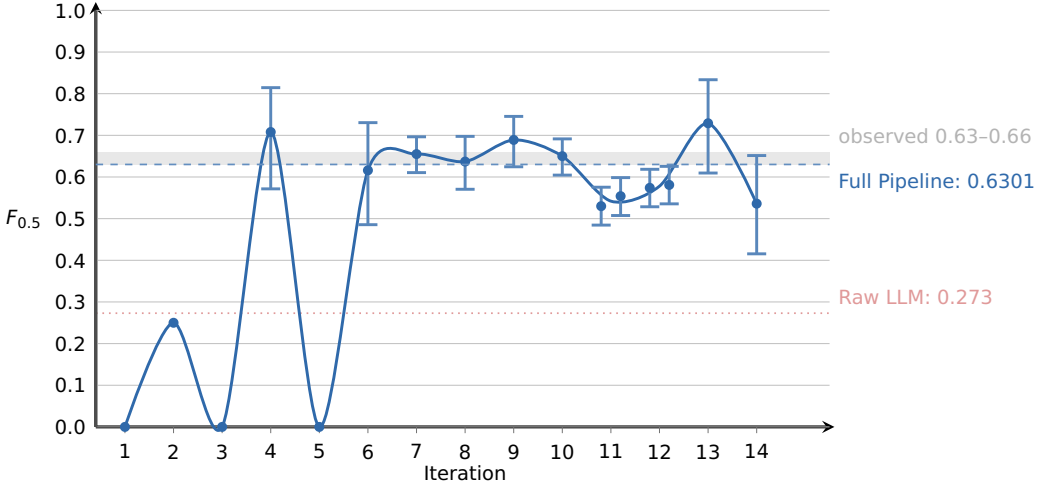
\FloatBarrier

On this reused development pool, the result is consistent with an
\emph{evidence-content plateau}: no tested composition rule recovers more signal
from the current M-check + U-check vocabulary. The trajectory in
Figure~\ref{fig:iterations} does not establish a performance ceiling beyond this
pool.

\subsection{Baselines and System-Level Ablation}
\label{sec:p1b-baselines}

\textbf{Conventional same-task baselines.} As a post hoc baseline check, we fit
two fixed linear classifiers using only fields available to the Raw LLM
baseline: category tags, short description, and founding year. A category-only
logistic model uses binary category indicators; a second logistic model uses
word-unigram/bigram TF--IDF over description, categories, and year. Both are
trained on the 198-row development draw. Their decision thresholds are selected
by five-fold stratified out-of-fold prediction on that draw, before one
evaluation on the row-disjoint 198-row validation draw. Implementations use
scikit-learn~\cite{pedregosa2011sklearn}. Table~\ref{tab:conventional-baselines}
reports the result.

\begin{center}
\begin{minipage}{\textwidth}
\centering
\captionof{table}{Conventional baselines on the row-disjoint validation draw.}
\label{tab:conventional-baselines}
\small
\renewcommand{\arraystretch}{1.12}
\begin{tabular}{lccc}
\toprule
System & Precision & Recall & $F_{0.5}$ [bootstrap 95\% CI] \\
\midrule
Category-only logistic & 0.284 & 0.396 & 0.3006 [0.191, 0.408] \\
Description + category TF--IDF logistic & 0.333 & 0.292 & 0.3241 [0.191, 0.453] \\
\textbf{Full Pipeline} & \textbf{0.529} & \textbf{0.563} & \textbf{0.5357 [0.412, 0.655]} \\
\bottomrule
\end{tabular}
\end{minipage}
\end{center}

Against the category-only and TF--IDF baselines, the pipeline's paired
$F_{0.5}$ differences are $+0.235$ [bootstrap 0.092, 0.377] and $+0.212$
[0.047, 0.378], respectively. These comparisons share the validation rows,
truth labels, and metric. They show that the pipeline exceeds two conventional
same-task classifiers on this draw; they do not establish deployment validity or
identify which pipeline component supplies the difference.

\textbf{Raw LLM baseline (A9).} A9 is the experiment identifier for the Raw LLM
baseline. To separate framework value from direct LLM inference, we
run a leading general-purpose LLM, with extended thinking disabled, on the same
396 rows. Inputs are anonymized and restricted to Crunchbase metadata (company
name, founding date, categories, and one-line tagline). The prompt requires the
model to use only the description, exclude post-founding information, and ignore
name recognition. This raw baseline scores
$F_{0.5}=0.2734$ (precision 0.875, recall 0.073): the LLM is extremely conservative
on idea-stage input---it recommends only 8 of 396 candidates, 7 correctly.

On VCBench's founder-profile task, GPT-4o leads the general-purpose model
baselines at $F_{0.5}=0.257$, while the task-trained Verifiable-RL system leads
the overall leaderboard at 0.366 at the time of access
\cite{vcbench,think_reason_learn}. VCBench differs from our study in data type, population
scope, and prediction target. These results are therefore not directly comparable
with ours. In particular, our numerically higher $F_{0.5}$ does not imply that
this pipeline outperforms systems on the VCBench leaderboard. We cite VCBench
because the two prediction problems are conceptually related and VCBench is one
of the most active communities studying this broader class of problems.

The apples-to-apples comparison in this study is instead between A9 and the full
pipeline: both use the same 396 clean rows, outcome labels, verdict schema, and
scoring rule under a shared leakage-controlled regime. The pipeline additionally
uses structured profiling, time-bounded retrieval, and multi-step reasoning. The
observed \textbf{$+0.357$} difference therefore
measures system-level framework lift over the Raw LLM baseline, but cannot be
assigned to any individual component. A row-aligned, outcome-stratified paired
bootstrap estimates this $F_{0.5}$ difference as $+0.3566$ (95\% CI
$0.1984$--$0.5312$); a paired prediction-swap randomization test gives
$p<0.00002$ (50{,}000 resamples). These tests quantify the end-to-end contrast
on the pooled benchmark; they complement the row-disjoint validation and
out-of-sample experiments reported above, but do not complete the larger external
scale validation or isolate any component's contribution.

\begin{figure*}[t]
\centering
\includegraphics[width=0.93\textwidth]{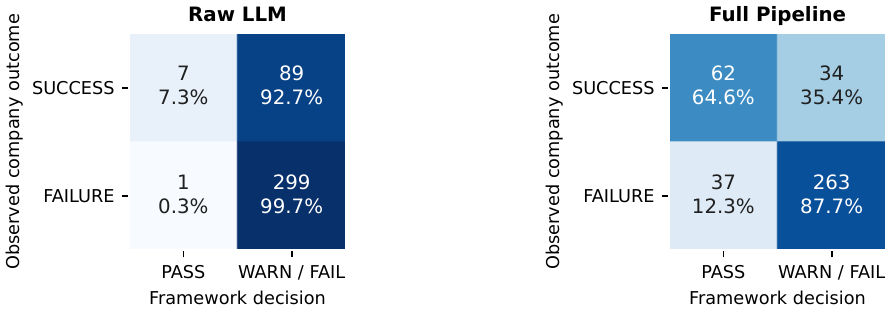}
\Description{Side-by-side row-normalized confusion matrices for Raw LLM and Full Pipeline, with counts and within-outcome percentages.}
\caption{Decision errors on the combined 396-row benchmark. Raw LLM
identifies 7 of 96 SUCCESS cases and produces one false positive; the full
pipeline identifies 62 SUCCESS cases while producing 37 false positives.
Percentages are normalized within the observed outcome rows.}
\label{fig:confusion-comparison}
\end{figure*}
\FloatBarrier

The error breakdown in Figure~\ref{fig:confusion-comparison} shows that Raw LLM
preserves very high specificity by nearly always rejecting, whereas the full
pipeline recovers substantially more SUCCESS cases at the cost of additional
false positives.

\textbf{Four-anchor ablation ladder.} We compare four system configurations along
an information ladder. Table~\ref{tab:ablation-ladder} reports the mean and
standard deviation across 10 seeds for the two randomization ablations; the other
configurations report single-run estimates.

\begin{center}
\begin{minipage}{\textwidth}
\centering
\captionof{table}{Four-anchor ablation ladder on the combined 396-row benchmark.}
\label{tab:ablation-ladder}
\renewcommand{\arraystretch}{1.15}
\setlength{\tabcolsep}{4pt}
\begin{tabular}{lcc}
\toprule
System & $F_{0.5}$ estimate & Role \\
\midrule
B1: uniform-random verdicts & 0.2565 $\pm$ 0.028 & zero-information floor \\
B2: random M/U + aggregate rule & 0.2553 $\pm$ 0.040 & aggregate on noise \\
Raw LLM (A9): metadata only & 0.2734 & LLM-native signal \\
\textbf{Full Pipeline v1.5a} & \textbf{0.6301} & evidence + aggregate \\
\bottomrule
\end{tabular}
\end{minipage}
\end{center}

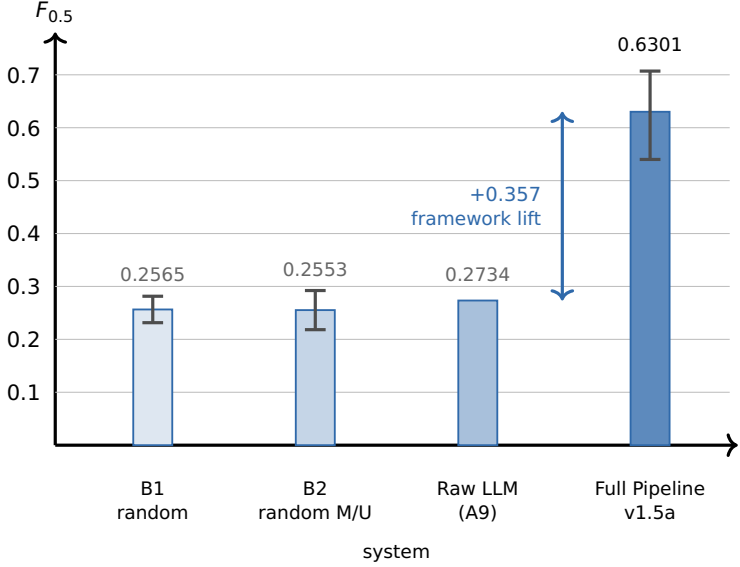
\begin{figure*}[t]
\centering
\begin{adjustbox}{max width=0.92\textwidth,center}
\begin{tikzpicture}[
  x=2.15cm, y=7.0cm,
  font=\figurebodyfont,
  every node/.style={font=\figurebodyfont},
  bar/.style={draw=reportBlue, line width=0.7pt},
  barfillone/.style={fill=reportBlue!16},
  barfilltwo/.style={fill=reportBlue!28},
  barfillthree/.style={fill=reportBlue!42},
  baraccent/.style={fill=reportBlue!78, draw=reportBlue},
  ebar/.style={|-|, thick, black!70},
]
\draw[->, thick] (0.55,0) -- (4.75,0);
\draw[->, thick] (0.55,0) -- (0.55,0.78) node[above] {$F_{0.5}$};
\node[below, font=\figurenotefont] at (2.65,-0.17) {system};

\foreach \y/\lab in {0.1/0.1,0.2/0.2,0.3/0.3,0.4/0.4,0.5/0.5,0.6/0.6,0.7/0.7}{
  \draw[black!25, very thin] (0.55,\y)--(4.7,\y);
  \node[left] at (0.52,\y) {\lab};
}

\draw[bar,barfillone] (1.03,0) rectangle (1.27,0.2565);
\draw[ebar] (1.15,0.2565-0.028) -- (1.15,0.2565+0.028);
\node[below, align=center, font=\figurenotefont] at (1.15,-0.05) {B1\\random};
\node[above, font=\figurenotefont, black!60] at (1.15,0.2565+0.035) {0.2565};

\draw[bar,barfilltwo] (2.03,0) rectangle (2.27,0.2553);
\draw[ebar] (2.15,0.2553-0.040) -- (2.15,0.2553+0.040);
\node[below, align=center, font=\figurenotefont] at (2.15,-0.05) {B2\\random M/U};
\node[above, font=\figurenotefont, black!60] at (2.15,0.2553+0.045) {0.2553};

\draw[bar,barfillthree] (3.03,0) rectangle (3.27,0.2734);
\node[below, align=center, font=\figurenotefont] at (3.15,-0.05) {Raw LLM\\(A9)};
\node[above, font=\figurenotefont, black!60] at (3.15,0.2734+0.018) {0.2734};

\draw[bar,baraccent] (4.09,0) rectangle (4.33,0.6301);
\draw[ebar] (4.21,0.537) -- (4.21,0.710);
\node[below, align=center, font=\figurenotefont] at (4.21,-0.05) {Full Pipeline\\v1.5a};
\node[above, yshift=3pt, font=\figurenotefont, black] at (4.21,0.710) {0.6301};

\draw[<->, thick, reportBlue] (3.67,0.2734) -- (3.67,0.6301);
\node[anchor=east, align=right, reportBlue, font=\figurenotefont]
  at (3.60,0.452) {$+0.357$\\framework lift};
\end{tikzpicture}
\end{adjustbox}
\Description{Bar chart comparing random, Raw LLM, and Full Pipeline F0.5 scores on the same development pool.}
\caption{Four-anchor ablation ladder on the same 396-row pool: uniform-random
verdicts (B1) and random M/U with a correct aggregate rule (B2) floor at
$\sim$0.256; Raw LLM (A9) on anonymized metadata adds negligible lift
($0.2734$); Full Pipeline v1.5a reaches $0.6301$
[bootstrap $0.537$--$0.710$]. The observed system-level difference from A9 is
$+0.357$ [paired bootstrap $0.198$--$0.531$; randomization $p<0.00002$], but
the changed inputs and reasoning setup prevent component attribution.}
\label{fig:ablation}
\end{figure*}

Figure~\ref{fig:ablation} visualizes the same four-anchor comparison.
Three descriptive comparisons follow. (i) Raw LLM exceeds its random
$F_{0.5}$ floor by only $0.017$; its rare
high-precision predictions do not move the metric materially because the metric penalizes its
extreme conservatism. (ii) The full pipeline exceeds the random M/U reference by
$0.375$. Because card richness, retrieval, reasoning mode, and orchestration
change together, this system-level difference cannot be assigned to the M/U
evidence layer alone. (iii) With realistically distributed noise as input, the aggregate
rule scores essentially the same as pure randomness ($\Delta=-0.001$):
deterministic composition does not create signal from noise.

\FloatBarrier

\subsection{Candidate Optimization Proof-of-Concept}
\label{sec:p1b-poc}

\begin{figure*}[t]
\centering
\includegraphics[width=0.96\textwidth]{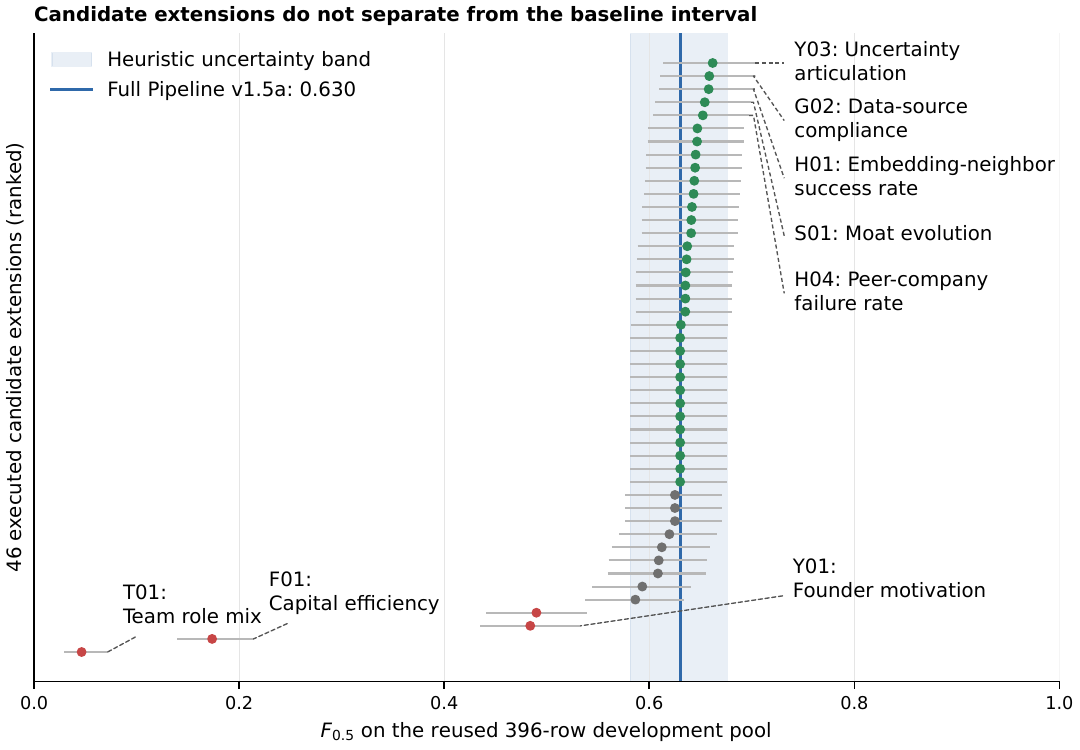}
\Description{Ranked forest plot of F0.5 estimates and recorded approximate intervals for 46 executed candidate extensions, against the baseline estimate and interval.}
\caption{Candidate-optimization landscape on the reused development pool. Points
are ranked $F_{0.5}$ estimates and horizontal lines are their recorded heuristic
uncertainty bands. The blue line and band mark Full Pipeline
v1.5a. None of the 46 executed extensions separates from the baseline interval;
the long negative tail shows
that plausible additions can materially degrade the decision rule.}
\label{fig:optimization-landscape}
\end{figure*}

We ran a systematic proof-of-concept sweep: 54 candidate
optimization ideas (new checks, prompt/dimension extensions, retrieval strategies,
and reasoning modes) were screened, of which 46 underwent full proof-of-concept
evaluation on the clean 396-row pool. Each was scored by combining a candidate signal with
the baseline M/U verdicts and re-optimizing the threshold. \textbf{No candidate
produced a clearly separated improvement on this reused development pool.} The best candidate reached
$F_{0.5}=0.6618$ [0.6138, 0.7066], overlapping the baseline $0.6301$
[0.5815, 0.6762] interval. Five of six retrieval-backed candidates collapsed to
exactly the baseline because the added signal had no useful effect at any tested
threshold. This negative result supports a development-pool
plateau interpretation, but cannot identify what would limit performance on an
untouched cohort.
The displayed candidate ranges are historical heuristic uncertainty bands from
the development audit, not formal confidence intervals for $F_{0.5}$ or paired
tests of improvement.
Figure~\ref{fig:optimization-landscape} summarizes the 46 executed extensions.

\FloatBarrier

\subsection{Limitations}
\label{sec:p1b-limits}

The benchmark has four local boundaries. \textbf{Validation scope.} The
independently drawn scale cohort remains partially executed. Its post-stratified
and composition-matched estimates align with the combined benchmark, but the
non-random completion process and untouched reserved split limit the strength of
the generalization claim. Completion was deferred for cost reasons: the first
1{,}024 completed reasoning runs cost about \$2{,}560.
\textbf{Input and outcome constructs.} Present-day
website text and category metadata need not reproduce a genuine founding-time
proposal; year-constrained retrieval cannot guarantee point-in-time web
availability; and SUCCESS/FAILURE measure funding or exit milestones rather
than profitability, product value, or long-run survival. \textbf{Imbalance and
uncertainty.} SUCCESS represents 4\% of the cohort. Historical per-iteration
ranges are approximate, whereas the combined headline uses a bootstrap interval.
\textbf{System-level attribution.} A9 and the full pipeline are apples-to-apples
at the evaluation level, but the treatment jointly changes card processing,
retrieval, reasoning mode, and orchestration. The comparison therefore estimates
system-level lift, not the isolated contribution or run-to-run variance of any
component. Sections~\ref{sec:general-limitations} and~\ref{sec:data} consolidate
the scale-validation and redistribution constraints.

\newcommand{\strategicClusterFigure}{%
\begin{figure*}[t]
\centering
\begin{adjustbox}{max width=\textwidth,center}
\begin{tikzpicture}[
  x=1.90cm, y=1.05cm,
  font=\figurebodyfont,
  every node/.style={font=\figurebodyfont},
  c1/.style={fill=reportGreen!28, draw=reportGreen, solid},
  c2/.style={fill=reportOrange!45, draw=reportOrange, dashed},
  c3/.style={fill=reportRed!40, draw=reportRed, densely dotted},
]
\draw[->, thick] (6.6,-0.15) -- (12.7,-0.15);
\draw[->, thick] (6.6,-0.15) -- (6.6,3.75);
\node[rotate=90] at (5.05,1.80) {independent replicability};
\node[font=\figurebodyfont] at (9.65,-0.84) {strategic scale (schematic)};
\draw[black!20, dashed] (6.6,1) -- (12.4,1);
\draw[black!20, dashed] (6.6,2) -- (12.4,2);
\draw[black!20, dashed] (6.6,3) -- (12.4,3);
\node[anchor=east, font=\figurenotefont, black!55] at (6.48,0) {none / near-none};
\node[anchor=east, font=\figurenotefont, black!55] at (6.48,1) {conditional};
\node[anchor=east, font=\figurenotefont, black!55] at (6.48,2) {replicable};
\node[anchor=east, font=\figurenotefont, black!55] at (6.48,3) {high};
\foreach \x/\lab in {7.0/{smaller},12.0/{larger}}{
  \node[below] at (\x,-0.26) {\lab};
}

\node[c2, circle, inner sep=0pt, minimum size=9mm, align=center] at (11.3,0.35) {\textbf{C1}\\{\tiny 8}};
\node[c2, circle, inner sep=0pt, minimum size=9mm, align=center] at (10.35,1.12) {\textbf{C2}\\{\tiny 4}};
\node[c2, circle, inner sep=0pt, minimum size=9mm, align=center] at (7.55,2.78) {\textbf{C3}\\{\tiny 3{+}1}};
\node[c3, circle, inner sep=0pt, minimum size=9mm, align=center] at (9.25,0.35) {\textbf{C4}\\{\tiny 6}};
\node[c2, circle, inner sep=0pt, minimum size=9mm, align=center] at (11.62,1.30) {\textbf{C5g}\\{\tiny 2}};
\node[c2, circle, inner sep=0pt, minimum size=9mm, align=center] at (8.95,2.15) {\textbf{C5l}\\{\tiny 2}};
\node[c3, circle, inner sep=0pt, minimum size=9mm, align=center] at (10.2,2.55) {\textbf{C6}\\{\tiny 7}};

\begin{scope}[yshift=-1.38cm]
\draw[c1] (6.7,0) rectangle (7.1,0.3);
\node[anchor=west, font=\figurenotefont] at (7.18,0.12) {profitable / near};
\draw[c2] (8.7,0) rectangle (9.1,0.3);
\node[anchor=west, font=\figurenotefont] at (9.18,0.12) {mixed / path-to-profit};
\draw[c3] (11.35,0) rectangle (11.75,0.3);
\node[anchor=west, font=\figurenotefont] at (11.83,0.12) {burning / subsidized};
\end{scope}

\node[font=\figurenotefont, black!70, anchor=north] at (9.5,-1.95) {C5g: giant-capex sub-tier; C5l: capital-light sub-tier};

\node[anchor=north west, align=left, text width=4.8cm,
      font=\figurenotefont] at (5.25,-2.50)
  {\textbf{C3: Consider}\\\textbf{Distribution-layer}\\
   For independent profitability;\\limited revenue ceiling.};
\node[anchor=north west, align=left, text width=4.8cm,
      font=\figurenotefont] at (8.05,-2.50)
  {\textbf{C5: Consider}\\\textbf{Frontier-model-owner}\\
   For capital-market upside;\\even the lean tier needs \$1--2B.};
\node[anchor=north west, align=left, text width=4.8cm,
      font=\figurenotefont] at (10.85,-2.50)
  {\textbf{C4: Avoid}\\\textbf{OSS-on-rented-GPU}\\
   As the default entry model;\\5 of 6 operators are burning.};
\end{tikzpicture}
\end{adjustbox}
\Description{A schematic map positions six strategic vendor clusters by qualitative strategic scale and independent replicability, with company counts inside markers. Three annotations distinguish distribution-layer entry for independent profitability, frontier-model ownership for capital-market upside with exceptional capital, and OSS-on-rented-GPU as a model to avoid by default. Horizontal positions are not comparable financial measurements.}
\caption{Six-cluster strategic map of the inference market. Numbers inside
markers report company counts; marker size has no quantitative meaning. Color
summarizes the profitability signal. Horizontal placement is schematic, drawing
on different measures---ARR, valuation, and financing scale---rather than a
common financial metric. The three annotations state conditional entry guidance,
not profitability forecasts. C3 includes three profitable or near-profitable
core vendors and one loss-making extension (SiliconFlow). C5 splits into giant-capex and capital-light
sub-tiers of two companies each.}
\label{fig:clusters}
\end{figure*}
}

\section{Industry Case Study: The AI-Inference Vendor Market}
\label{sec:part1-casestudy}

This section reports a systematic, bottom-up study of inference operators: a
29-vendor analytical snapshot collected in June 2026\footnote{Some vendors have since undergone
material changes in operating status. The evidence and conclusions in this case
study are bounded by the June 2026 collection cutoff and should be interpreted as
claims about that snapshot, not vendors' subsequent condition.}
comprising a seven-archetype taxonomy, a profitability crosstab, and a
six-cluster strategic map. The released
\texttt{evaluate-proposal} workflow does not classify candidates into these
archetypes.
The study follows the benchmark's evidence discipline
(Section~\ref{sec:p1b-leakage}): every number is linked to a dated public source
and an evidence-quality rating; undisclosed values remain missing.

The strategic takeaway is conditional on the founder's objective:
\textbf{consider Distribution-layer for independent profitability}, with a
limited revenue ceiling; \textbf{consider Frontier-model-owner for capital-market
upside}, only with access to exceptional capital; and \textbf{avoid
OSS-on-rented-GPU as the default independent-entry model}, given its weak
observed economics. Figure~\ref{fig:clusters} relates these conclusions to the
strategic map.

\subsection{Scope, Card Protocol, and Source Discipline}
\label{sec:p1cs-scope}

\textbf{Scope.} The purposive 29-vendor sample is not a market census. It begins
with 21 API service providers from a prior workload-specification taxonomy,
adds Hugging Face and seven omissions identified by a same-model price audit,
and excludes a later CoreWeave reference card from the frozen crosstab.
Cards were assembled in June 2026 from the MIT-licensed OpenSporks Crunchbase
free database snapshot downloaded from Hugging Face, supplemented with public
information available at the time.

\textbf{Card protocol.} Each vendor is documented in a bounded-length \emph{vendor
card} following a five-dimensional schema---revenue model, customer segmentation,
cost structure, differentiation and moat, and strategic vulnerabilities. The
prediction pipeline reuses these five dimensions in its forward-looking candidate
card (Section~\ref{sec:p1m-card}): proposal claims replace observed facts, and
per-dimension confidence becomes a plausibility flag.

\textbf{Source discipline.} Staged public-web search, full-page extraction, and
refresh rounds covered filings, company disclosures, product and pricing pages,
and other public evidence. Collection targeted at least three independent sources
per card and retained URLs or screenshots. Claim-level ratings distinguish
audited disclosure, reported figures, estimates, and undisclosed fields.
Across all 30 cards, including the CoreWeave reference, 22 carry explicit
\texttt{data-thin} flags totaling more than 100 points of missing disclosure.

\begin{figure*}[t]
\centering
\begin{adjustbox}{max width=\textwidth,center}
\begin{tikzpicture}[
  x=1.0cm, y=0.5cm,
  font=\figurebodyfont,
  every node/.style={font=\figurebodyfont},
  prof/.style={fill=reportGreen!28, draw=reportGreen, report positive fill},
  near/.style={fill=reportOrange!55, draw=reportOrange, report intermediate fill},
  burn/.style={fill=reportRed!55, draw=reportRed, report negative fill},
]
\draw[prof] (0,6) rectangle (5,6.6);
\draw[near] (0,5) rectangle (3,5.6);
\draw[burn] (3,5) rectangle (5,5.6);
\draw[near] (0,4) rectangle (1,4.6);
\draw[burn] (1,4) rectangle (2,4.6);
\draw[prof] (0,3) rectangle (2,3.6);
\draw[near] (2,3) rectangle (3,3.6);
\draw[near] (0,2) rectangle (1,2.6);
\draw[burn] (1,2) rectangle (6,2.6);
\draw[prof] (0,1) rectangle (1,1.6);
\draw[near] (1,1) rectangle (3,1.6);
\draw[burn] (0,0) rectangle (5,0.6);

\node[anchor=west, font=\figurenotefont] at (5.25,6.28) {Hyperscaler-bundle (5)};
\node[anchor=west, font=\figurenotefont] at (5.25,5.28) {Frontier-model-owner (5)};
\node[anchor=west, font=\figurenotefont] at (2.25,4.28) {Hardware-vertical (2)};
\node[anchor=west, font=\figurenotefont] at (3.25,3.28) {Enterprise/Rental (3)};
\node[anchor=west, font=\figurenotefont] at (6.25,2.28) {OSS-on-rented-GPU (6)};
\node[anchor=west, font=\figurenotefont] at (3.25,1.28) {Distribution-layer (3)};
\node[anchor=west, font=\figurenotefont] at (5.25,0.28) {CN-private (5)};

\draw[->, thick] (0,-0.6) -- (10.2,-0.6);
\node[below] at (5.1,-1.3) {Number of vendors};
\foreach \x in {0,...,10}{
  \draw[black!55] (\x,-0.6)--(\x,-0.68);
  \node[below] at (\x,-0.7) {\x};
}

\begin{scope}[yshift=-1.9cm,font=\figurenotefont]
\draw[prof] (0,0) rectangle (0.55,0.3); \node[anchor=west] at (0.7,0.1) {profitable};
\draw[near] (2.8,0) rectangle (3.35,0.3); \node[anchor=west] at (3.5,0.1) {near-profitable};
\draw[burn] (6.35,0) rectangle (6.9,0.3); \node[anchor=west] at (7.05,0.1) {burning / subsidized};
\end{scope}
\end{tikzpicture}
\end{adjustbox}
\Description{Stacked bars compare profitability classifications across seven inference-market business-model archetypes.}
\caption{Archetype $\times$ profitability composition across the 29-vendor scope.
Bars show the number of vendors per profitability signal (green = profitable;
orange = near-profitable; red = burning / subsidized). Profitability is
concentrated in hyperscaler parents, enterprise/rental operators, and the
distribution layer; the ``lose money but raise capital'' profile concentrates in
frontier model owners and OSS-on-rented-GPU operators.}
\label{fig:archetype-profit}
\end{figure*}
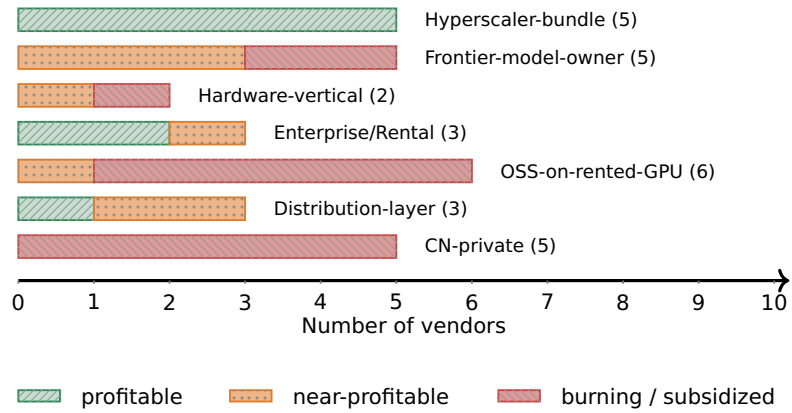
\FloatBarrier

\subsection{Seven Archetypes and the Profit Crosstab}
\label{sec:p1cs-arch}

The cards were first grouped \emph{bottom-up} into business-model archetypes by
common revenue and cost structure, not by market share or brand. The resulting
seven-archetype taxonomy is a descriptive coding scheme for this market study,
not an input to the released proposal evaluator. Figure~\ref{fig:archetype-profit}
shows the comparison that motivates the category-by-category definitions below;
Table~\ref{tab:archetype-profit} subsequently retains the exact counts.

\begin{itemize}[label=\textbullet]
  \item \textbf{Hyperscaler-bundle} --- inference sold as one line on a hyperscale cloud
        bill (Bedrock, Azure OpenAI/Foundry, Vertex, Bailian, Qianfan);
  \item \textbf{Frontier-model-owner} --- operators who train and own their frontier or
        domain models as core IP (Anthropic, OpenAI, xAI; capital-light variants
        Mistral, Cohere; plus CN frontier labs);
  \item \textbf{Hardware-vertical} --- chip-and-system manufacturers who run an inference
        cloud as a secondary product line (Cerebras, Groq);
  \item \textbf{Enterprise/Rental} --- committed-capacity GPU lessors and data-AI
        platforms (Lambda, OVHcloud, later CoreWeave; Databricks as the
        data-platform anomaly);
  \item \textbf{OSS-on-rented-GPU} --- rented-GPU hosts of open-weight models priced
        per-token or per-second (Together, Fireworks, Modal, Baseten, RunPod,
        DeepInfra);
  \item \textbf{Distribution-layer} --- aggregators and model hubs living primarily on
        routing, credit fees, and network effects (OpenRouter, Hugging Face, and
        Replicate);
  \item \textbf{CN-private} --- a geographic and regulatory bucket, rather than a pure business
        model, for China-market operators shaped by regulation and data
        residency (Volcengine Ark, Zhipu, Moonshot, DeepSeek, SiliconFlow).
\end{itemize}

Here, OSS denotes open-source software.
The first six labels describe operating economics, whereas \emph{CN-private}
describes market jurisdiction and can overlap them conceptually. For the crosstab,
each vendor is assigned once by the study's frozen precedence rule; the taxonomy
should therefore be read as a practical coding scheme, not mutually exclusive
latent classes.

\textbf{The profit crosstab} (Table~\ref{tab:archetype-profit}) reports the
study's coded profitability signal for every in-scope vendor. Profitability is coded
\emph{independently of
capital-market success}: an operator that burns cash but commands a high valuation
is counted as burning, not as profitable; the two dimensions are documented
separately throughout the study.

\begin{table}[t]
\centering
\caption{Archetype $\times$ profitability crosstab across the 29-vendor scope
(June 2026 snapshot). ``Near'' = path to profitability disclosed or
capital-efficient with likely positive unit economics; ``subsidy'' = losses carried
by a parent or state anchor.}
\label{tab:archetype-profit}
\begin{tabular}{@{}l*{4}{>{\centering\arraybackslash}p{0.14\linewidth}}@{}}
\toprule
Archetype & $n$ & Profitable & Near & Burning / subsidy \\
\midrule
Hyperscaler-bundle & 5 & 5 & 0 & 0 \\
Frontier-model-owner & 5 & 0 & 3 & 2 \\
Hardware-vertical & 2 & 0 & 1 & 1 \\
Enterprise/Rental & 3 & 2 & 1 & 0 \\
OSS-on-rented-GPU & 6 & 0 & 1 & 5 \\
Distribution-layer & 3 & 1 & 2 & 0 \\
CN-private & 5 & 0 & 0 & 3 $+$ 2 \\
\midrule
Total & 29 & 8 (28\%) & 8 (28\%) & 11 $+$ 2 \\
\bottomrule
\end{tabular}
\end{table}

All percentages and counts below describe this purposive sample. For bundled
products, the profitability code can reflect parent or cloud-segment economics
rather than standalone inference profitability; estimated values are not audited
accounting measures.

Three conclusions drive the framework design. \textbf{(i) Profitability is
concentrated where the operator does not bear independent compute economics.} The
eight profitable operators comprise hyperscaler parents (whose inference revenue is one
line in an already-profitable cloud segment), a data-AI platform, an EU sovereign
cloud, or a post-acquisition distribution operator --- none is an independent,
inference-core company. \textbf{(ii) The ``lose money but raise capital'' profile is
concentrated in two archetypes}: frontier model owners (capital markets
value scarce model IP despite projected 2026--2029 burn) and OSS-on-rented-GPU
operators (5/6 burning, no pending IPOs, and intense price competition in
inference-stack kernels).
\textbf{(iii) A corollary for interpretation}: the OSS-on-rented pattern warns
analysts to scrutinize unit economics, while distribution-layer cases motivate
separate attention to market ceiling. These are case-study observations rather
than archetype priors or inputs to the released M/U checks.

\subsection{Six-Cluster Strategic Map}
\label{sec:p1cs-clusters}

A second, \emph{top-down} analyst grouping assigns the same vendors to potentially
overlapping strategic clusters based on capital structure, customer base, and
replicability. Its six clusters are the study's answer to ``which kind of business
is worth doing'' (Table~\ref{tab:strategic-clusters}):

\ifdefined\TenPtClose
\begin{figure*}[t]
\centering
\begin{adjustbox}{max width=\textwidth,center}
\begin{tikzpicture}[
  x=1.90cm, y=1.05cm,
  font=\figurebodyfont,
  every node/.style={font=\figurebodyfont},
  c1/.style={fill=reportGreen!28, draw=reportGreen, solid},
  c2/.style={fill=reportOrange!45, draw=reportOrange, dashed},
  c3/.style={fill=reportRed!40, draw=reportRed, densely dotted},
]
\draw[->, thick] (6.6,-0.15) -- (12.7,-0.15);
\draw[->, thick] (6.6,-0.15) -- (6.6,3.75);
\node[rotate=90] at (5.05,1.80) {independent replicability};
\node[font=\figurebodyfont] at (9.65,-0.84) {strategic scale (schematic)};
\draw[black!20, dashed] (6.6,1) -- (12.4,1);
\draw[black!20, dashed] (6.6,2) -- (12.4,2);
\draw[black!20, dashed] (6.6,3) -- (12.4,3);
\node[anchor=east, font=\figurenotefont, black!55] at (6.48,0) {none / near-none};
\node[anchor=east, font=\figurenotefont, black!55] at (6.48,1) {conditional};
\node[anchor=east, font=\figurenotefont, black!55] at (6.48,2) {replicable};
\node[anchor=east, font=\figurenotefont, black!55] at (6.48,3) {high};
\foreach \x/\lab in {7.0/{smaller},12.0/{larger}}{
  \node[below] at (\x,-0.26) {\lab};
}

\node[c2, circle, inner sep=0pt, minimum size=9mm, align=center] at (11.3,0.35) {\textbf{C1}\\{\tiny 8}};
\node[c2, circle, inner sep=0pt, minimum size=9mm, align=center] at (10.35,1.12) {\textbf{C2}\\{\tiny 4}};
\node[c2, circle, inner sep=0pt, minimum size=9mm, align=center] at (7.55,2.78) {\textbf{C3}\\{\tiny 3{+}1}};
\node[c3, circle, inner sep=0pt, minimum size=9mm, align=center] at (9.25,0.35) {\textbf{C4}\\{\tiny 6}};
\node[c2, circle, inner sep=0pt, minimum size=9mm, align=center] at (11.62,1.30) {\textbf{C5g}\\{\tiny 2}};
\node[c2, circle, inner sep=0pt, minimum size=9mm, align=center] at (8.95,2.15) {\textbf{C5l}\\{\tiny 2}};
\node[c3, circle, inner sep=0pt, minimum size=9mm, align=center] at (10.2,2.55) {\textbf{C6}\\{\tiny 7}};

\begin{scope}[yshift=-1.38cm]
\draw[c1] (6.7,0) rectangle (7.1,0.3);
\node[anchor=west, font=\figurenotefont] at (7.18,0.12) {profitable / near};
\draw[c2] (8.7,0) rectangle (9.1,0.3);
\node[anchor=west, font=\figurenotefont] at (9.18,0.12) {mixed / path-to-profit};
\draw[c3] (11.35,0) rectangle (11.75,0.3);
\node[anchor=west, font=\figurenotefont] at (11.83,0.12) {burning / subsidized};
\end{scope}

\node[font=\figurenotefont, black!70, anchor=north] at (9.5,-1.95) {C5g: giant-capex sub-tier; C5l: capital-light sub-tier};

\node[anchor=north west, align=left, text width=4.8cm,
      font=\figurenotefont] at (5.25,-2.50)
  {\textbf{C3: Consider}\\\textbf{Distribution-layer}\\
   For independent profitability;\\limited revenue ceiling.};
\node[anchor=north west, align=left, text width=4.8cm,
      font=\figurenotefont] at (8.05,-2.50)
  {\textbf{C5: Consider}\\\textbf{Frontier-model-owner}\\
   For capital-market upside;\\even the lean tier needs \$1--2B.};
\node[anchor=north west, align=left, text width=4.8cm,
      font=\figurenotefont] at (10.85,-2.50)
  {\textbf{C4: Avoid}\\\textbf{OSS-on-rented-GPU}\\
   As the default entry model;\\5 of 6 operators are burning.};
\end{tikzpicture}
\end{adjustbox}
\Description{A schematic map positions six strategic vendor clusters by qualitative strategic scale and independent replicability, with company counts inside markers. Three annotations distinguish distribution-layer entry for independent profitability, frontier-model ownership for capital-market upside with exceptional capital, and OSS-on-rented-GPU as a model to avoid by default. Horizontal positions are not comparable financial measurements.}
\caption{Six-cluster strategic map of the inference market. Numbers inside
markers report company counts; marker size has no quantitative meaning. Color
summarizes the profitability signal. Horizontal placement is schematic, drawing
on different measures---ARR, valuation, and financing scale---rather than a
common financial metric. The three annotations state conditional entry guidance,
not profitability forecasts. C3 includes three profitable or near-profitable
core vendors and one loss-making extension (SiliconFlow). C5 splits into giant-capex and capital-light
sub-tiers of two companies each.}
\label{fig:clusters}
\end{figure*}

\fi

\begin{center}
\begin{minipage}{\textwidth}
\centering
\captionof{table}{Strategic cross-tags in the 29-vendor AI-inference case study.}
\label{tab:strategic-clusters}
\small
\renewcommand{\arraystretch}{1.2}
\setlength{\tabcolsep}{3.2pt}
\begin{tabular}{l>{\raggedright\arraybackslash}p{0.22\textwidth}>{\raggedright\arraybackslash}p{0.16\textwidth}>{\raggedright\arraybackslash}p{0.24\textwidth}}
\toprule
Cluster ($n$) & Operator examples & Profit state & Independent-company replicability \\
\midrule
C1 Hyperscaler-cluster (8) & Bedrock, Foundry, Vertex, Databricks & 5 profitable; 3 parent-subsidized & none (needs pre-existing cloud) \\
C2 Sovereign-AI/hardware (4) & Cerebras, Groq, Lambda, OVHcloud & mixed & only with sovereign anchor \\
C3 Long-tail distribution (3+1) & OpenRouter, Hugging Face, Replicate; SiliconFlow (extension) & 3 profitable/near; 1 burning & high for core three; their combined estimated ARR $< \$300$M \\
C4 OSS-on-rented (6) & Together, Fireworks, Modal, RunPod & 5 burning; 1 near-profitable & trap; exit is acquisition \\
C5 Frontier capital tier (4) & Anthropic, OpenAI, Mistral, Cohere & path-to-profit 2026--2028 & giant sub-tier not; lean \$1--2B \\
C6 CN-based (7) & Volcengine, Zhipu, DeepSeek, Moonshot & mostly subsidized & CN companies only \\
\bottomrule
\end{tabular}
\end{minipage}
\end{center}

Because these clusters are strategic cross-tags rather than a partition, their
counts do not sum to 29.
Figure~\ref{fig:clusters} schematically maps the six cross-tags by strategic scale and
independent-company replicability.

\ifdefined\TenPtClose\else
\begin{figure*}[t]
\centering
\begin{adjustbox}{max width=\textwidth,center}
\begin{tikzpicture}[
  x=1.90cm, y=1.05cm,
  font=\figurebodyfont,
  every node/.style={font=\figurebodyfont},
  c1/.style={fill=reportGreen!28, draw=reportGreen, solid},
  c2/.style={fill=reportOrange!45, draw=reportOrange, dashed},
  c3/.style={fill=reportRed!40, draw=reportRed, densely dotted},
]
\draw[->, thick] (6.6,-0.15) -- (12.7,-0.15);
\draw[->, thick] (6.6,-0.15) -- (6.6,3.75);
\node[rotate=90] at (5.05,1.80) {independent replicability};
\node[font=\figurebodyfont] at (9.65,-0.84) {strategic scale (schematic)};
\draw[black!20, dashed] (6.6,1) -- (12.4,1);
\draw[black!20, dashed] (6.6,2) -- (12.4,2);
\draw[black!20, dashed] (6.6,3) -- (12.4,3);
\node[anchor=east, font=\figurenotefont, black!55] at (6.48,0) {none / near-none};
\node[anchor=east, font=\figurenotefont, black!55] at (6.48,1) {conditional};
\node[anchor=east, font=\figurenotefont, black!55] at (6.48,2) {replicable};
\node[anchor=east, font=\figurenotefont, black!55] at (6.48,3) {high};
\foreach \x/\lab in {7.0/{smaller},12.0/{larger}}{
  \node[below] at (\x,-0.26) {\lab};
}

\node[c2, circle, inner sep=0pt, minimum size=9mm, align=center] at (11.3,0.35) {\textbf{C1}\\{\tiny 8}};
\node[c2, circle, inner sep=0pt, minimum size=9mm, align=center] at (10.35,1.12) {\textbf{C2}\\{\tiny 4}};
\node[c2, circle, inner sep=0pt, minimum size=9mm, align=center] at (7.55,2.78) {\textbf{C3}\\{\tiny 3{+}1}};
\node[c3, circle, inner sep=0pt, minimum size=9mm, align=center] at (9.25,0.35) {\textbf{C4}\\{\tiny 6}};
\node[c2, circle, inner sep=0pt, minimum size=9mm, align=center] at (11.62,1.30) {\textbf{C5g}\\{\tiny 2}};
\node[c2, circle, inner sep=0pt, minimum size=9mm, align=center] at (8.95,2.15) {\textbf{C5l}\\{\tiny 2}};
\node[c3, circle, inner sep=0pt, minimum size=9mm, align=center] at (10.2,2.55) {\textbf{C6}\\{\tiny 7}};

\begin{scope}[yshift=-1.38cm]
\draw[c1] (6.7,0) rectangle (7.1,0.3);
\node[anchor=west, font=\figurenotefont] at (7.18,0.12) {profitable / near};
\draw[c2] (8.7,0) rectangle (9.1,0.3);
\node[anchor=west, font=\figurenotefont] at (9.18,0.12) {mixed / path-to-profit};
\draw[c3] (11.35,0) rectangle (11.75,0.3);
\node[anchor=west, font=\figurenotefont] at (11.83,0.12) {burning / subsidized};
\end{scope}

\node[font=\figurenotefont, black!70, anchor=north] at (9.5,-1.95) {C5g: giant-capex sub-tier; C5l: capital-light sub-tier};

\node[anchor=north west, align=left, text width=4.8cm,
      font=\figurenotefont] at (5.25,-2.50)
  {\textbf{C3: Consider}\\\textbf{Distribution-layer}\\
   For independent profitability;\\limited revenue ceiling.};
\node[anchor=north west, align=left, text width=4.8cm,
      font=\figurenotefont] at (8.05,-2.50)
  {\textbf{C5: Consider}\\\textbf{Frontier-model-owner}\\
   For capital-market upside;\\even the lean tier needs \$1--2B.};
\node[anchor=north west, align=left, text width=4.8cm,
      font=\figurenotefont] at (10.85,-2.50)
  {\textbf{C4: Avoid}\\\textbf{OSS-on-rented-GPU}\\
   As the default entry model;\\5 of 6 operators are burning.};
\end{tikzpicture}
\end{adjustbox}
\Description{A schematic map positions six strategic vendor clusters by qualitative strategic scale and independent replicability, with company counts inside markers. Three annotations distinguish distribution-layer entry for independent profitability, frontier-model ownership for capital-market upside with exceptional capital, and OSS-on-rented-GPU as a model to avoid by default. Horizontal positions are not comparable financial measurements.}
\caption{Six-cluster strategic map of the inference market. Numbers inside
markers report company counts; marker size has no quantitative meaning. Color
summarizes the profitability signal. Horizontal placement is schematic, drawing
on different measures---ARR, valuation, and financing scale---rather than a
common financial metric. The three annotations state conditional entry guidance,
not profitability forecasts. C3 includes three profitable or near-profitable
core vendors and one loss-making extension (SiliconFlow). C5 splits into giant-capex and capital-light
sub-tiers of two companies each.}
\label{fig:clusters}
\end{figure*}
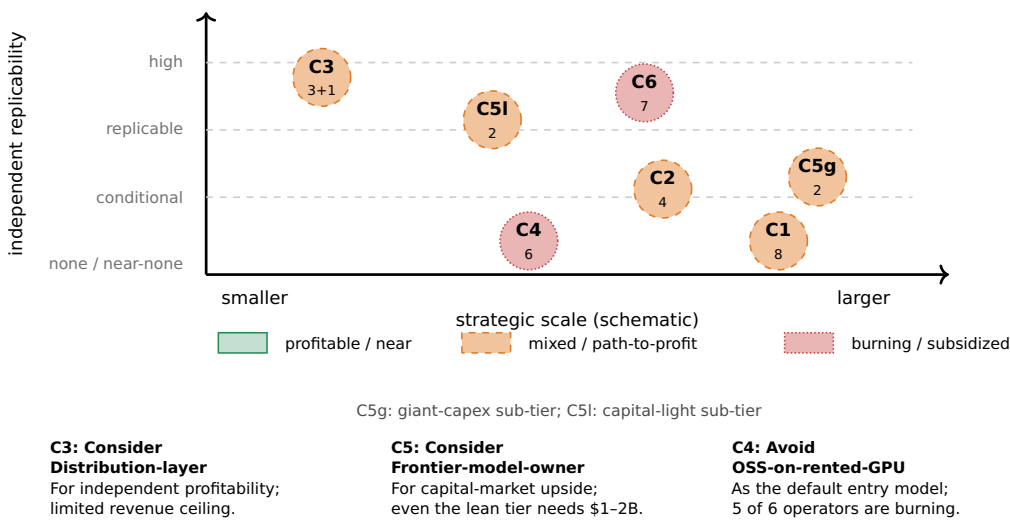

\fi
\FloatBarrier

The map is an analyst-coded visualization of the selected cards, not an estimated
population model. Its annual recurring revenue (ARR), valuation-multiple, and starting-capital ranges combine
reported figures with card-level estimates and should be read as hypotheses.

Three observations support the entry guidance stated at the start of this section.
\textbf{Distribution-layer (C3): independent profitability.} All three vendors
coded as Distribution-layer are profitable or near-profitable, with network
effects and distribution economics supporting this path. However,
their combined estimated ARR is below \$300M, and hyperscaler zero-fee routing threatens
the revenue ceiling.
\textbf{Frontier-model-owner (C5): capital-market upside.} The cards report
30--65$\times$ revenue multiples despite continuing losses or uncertain paths to
profitability. Even the capital-light sub-tier (Mistral and Cohere) requires an
estimated \$1--2B in starting capital: this is a capital-intensive opportunity,
not an alternative low-cost route to profitability.
\textbf{OSS-on-rented-GPU (C4): avoid as the default entry model.} Five of six
operators are coded as burning and one as near-profitable; none has a pending IPO
or a sovereign customer base. RunPod's near-profitable position is attributed to
community go-to-market execution rather than an inference-stack-IP narrative,
underscoring the need for an advantage beyond serving open-weight models.

\subsection{Cross-Cluster Patterns and Trajectories}
\label{sec:p1cs-patterns}

Six cross-cluster patterns and five 12--24-month trajectories complete the picture.
Four patterns are structurally important. First, \emph{sovereignty-driven
restructuring} moves roughly one-third of market volume through state capital,
sovereign anchor customers, or public R\&D funding; these vendors require strategic-
rather than unit-economics-based evaluation. Second, \emph{capital bifurcation}
divides each archetype into giant-capex and capital-light paths with a roughly
50$\times$ difference in burn rate. Third, a \emph{race-to-the-bottom trap} arises
because inference kernels diffuse quickly, new chips reset the field, and
aggregators arbitrage prices. Fourth, the \emph{long-tail profitability paradox}
makes easy profitability and scalability difficult to achieve together.

Scenario trajectories through 2027--2028 place the hardware-vertical segment at an inflection,
frontier owners in a possible dual-tier maturity phase, distribution under pressure
from hyperscaler routing, and OSS-on-rented under consolidation pressure. These are
analyst scenarios derived from the June-2026 snapshot, not verified future events or
probability-calibrated forecasts.

\subsection{Limitations}
\label{sec:p1cs-limits}

This case study has five local boundaries. \textbf{Cross-sectional, not
predictive.} The crosstab is frozen at June 2026. Vendors labeled ``burning,''
including Anthropic and Cerebras, have disclosed paths to profitability or an
IPO; these are trajectory descriptions, not forecasts. \textbf{Disclosure
asymmetry.} Profitability assignments inherit each card's data-quality rating,
and undisclosed figures remain unestimated; Section~\ref{sec:p1cs-scope}
quantifies the \texttt{data-thin} cases. \textbf{Conditional decision guidance.}
The entry guidance depends on founder objectives, capital access, and the observed
snapshot; it is not an automated recommendation or a guarantee of business
success. The archetypes are not runtime fields or predictor
inputs. \textbf{Taxonomy boundary.} The \emph{CN-private} code mixes jurisdiction
with business model, some vendors span multiple archetypes, and single-label
counts depend on the frozen assignment rule. \textbf{Sampling boundary.} The
purposive sample omits unknown and less-visible operators, so its percentages are
neither market shares nor population frequencies. Broader time-sensitivity and
release constraints are consolidated in Sections~\ref{sec:general-limitations}
and~\ref{sec:data}.

\par\noindent\begin{minipage}{\textwidth}
\researchpart{Two}{Post-Founding Decision Assistance}

After securing funding, startup founders face two linked decisions:
\textbf{(1) Which external operating actions should they pursue to advance
toward the next financing milestone? (2) Which investors or investment
institutions should they proactively approach, given their company's objectives
and those investors' observed behavior?} Part~Two develops a public-evidence
event-chain framework to inform these decisions (Figure~\ref{fig:part2-architecture}).

\begin{center}
\centering
\includegraphics[width=\textwidth]{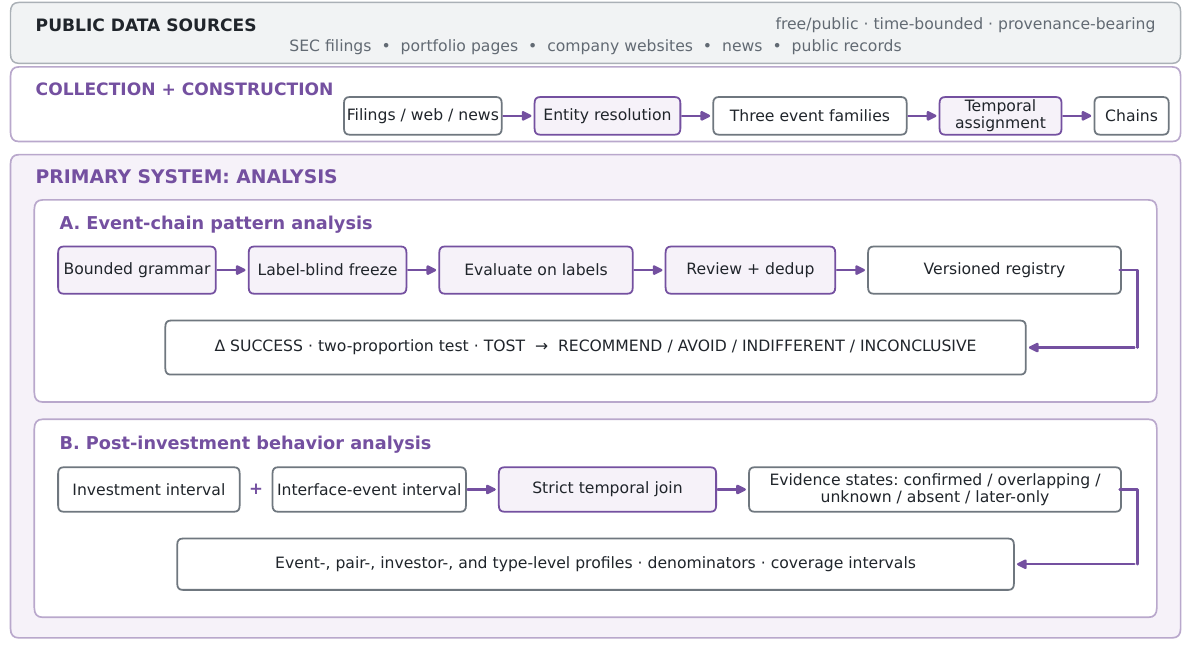}
\Description{Part Two architecture: public-evidence collection, event-chain construction, label-blind pattern analysis, and retrospective post-investment behavior analysis.}
\ifdefined\captionsetup\captionsetup{hypcap=false}\fi
\captionof{figure}{Part~Two architecture: event collection, chain construction, and descriptive analysis.}
\label{fig:part2-architecture}
\end{center}
\end{minipage}\par

Sections~\ref{sec:part2-data}--\ref{sec:part2-collection} describe the framework's
data design and construction; Sections~\ref{sec:part2-patterns}--\ref{sec:part2-behavior}
analyze operating-action patterns and investor behavior, respectively.

\section{Data Structure and Domain Scoping}
\label{sec:part2-data}

Answering the two founder questions requires linking what a company does,
which investors participate, and which financing or exit milestones follow.
Part~Two represents this evidence as an investor--company event graph, from which
company event chains and investor histories can be reconstructed.
Stable identifiers, explicit date intervals, and source-level provenance make
the structure auditable and reusable beyond the chip-company application.

\subsection{Two Research Entry Points}
\label{sec:p2d-entry}

The structure supports two domain-scoping strategies. A \emph{company-first}
study defines the industry and target company subset before targeted collection
of the relevant investors and events. The chip-company application uses this
route, drawing its scope from the broader entity and portfolio infrastructure
described in Section~\ref{sec:part2-collection}.
An \emph{investor-first} study instead begins with selected investors'
complete profiles and portfolios, then filters or compares companies by industry.
In either route, the target domain must be fixed before downstream event-pattern
evaluation; changing it later changes both coverage and the analytical population.

\subsection{Graph and Event Semantics}
\label{sec:p2d-structure}

Investor and company nodes are connected by portfolio and investment relations.
The core analytical object is the \emph{event chain}: one chain per boundary
outcome, spanning from founding (or the previous boundary) to that outcome, with
dated intermediate events. Three event families distinguish the milestone from
the operating developments and investor actions around it:

\begin{itemize}
  \item \textbf{Chain-boundary events} (\texttt{funding\_events}): funding rounds,
        failed or withdrawn financings, IPOs, acquisitions, and closures;
  \item \textbf{Exposure events} (\texttt{exposure\_events}):
        product launches, supply-chain moves, customer changes, regulatory and
        export-control actions, executive changes, litigation;
  \item \textbf{Interface events} (\texttt{interface\_\allowbreak events}):
        funding participation, board changes, major-holder changes, secondary
        transactions, strategic investments, and activist actions.
\end{itemize}

The graph separates row grain from analytical grain. Source rows retain the
investor, company, event, date interval, URL, and provenance needed for audit.
Entity resolution maps those rows to stable investor and company identities.
Funding, exposure, and interface records remain separate typed event families,
and chain materialization groups participant-grain funding rows into one logical
chain-boundary node. This prevents co-investor rows from becoming duplicate timeline
events or duplicated capital totals.

\section{Data Collection and Event-Chain Construction}
\label{sec:part2-collection}

We instantiate this structure from freely available public sources. The
production pipeline retrieves, caches, resolves, reviews, and versions each layer
before chain construction. The broad investor and company infrastructure supports
domain selection; targeted event collection then reconstructs the chip-company
histories analyzed here. The broad-universe counts below are therefore not the
chip analysis sample sizes. Paid-database rows are not included in the released tables.

\subsection{Scope and Sources}
\label{sec:p2d-scope}

The shared infrastructure was developed through a US institutional-VC pilot and
a global scale-up covering nine investor types. Sources include
SEC EDGAR Form D and 13D/13G filings~\cite{sec_edgar}, a public
findfunding.vc API, Wikidata SPARQL~\cite{wikidata} (CC0), corporate portfolio pages
(scraped with per-page CSS selector configurations), public news via a search API
\cite{tavily}, S-1 prospectuses, analyst 13F holdings, and national-fund disclosures
(GPIF Excel, NBIM/CPP Investments PDFs, CalSTRS HTML, pension disclosures).
Retrieval scripts are versioned and idempotent, and cache raw responses by date.

\subsection{Investor Entity Reconstruction}
\label{sec:p2d-entities}

\textbf{Pilot US-VC layer.} Four sources are merged with union--find deduplication and
canonical-value selection by confidence: findfunding.vc (1{,}369 firms),
Wikidata (110), a manually curated seed list (91), and a retrospective URL
verification pass (64 corrections). The result is \emph{1{,}446 canonical US
institutional investors} with 11{,}891 long-format provenance rows recording each
field's source and assigned confidence. Every canonical investor
carries a stable ID.

\textbf{Scale-up global layer.} The universe expands to a nine-type taxonomy
based on industry convention rather than mutually exclusive academic categories: hedge
fund, private equity, sovereign wealth, pension fund, asset manager, investment
bank, VC, corporate VC (CVC), and angel. Tier assignment within each type is determined for the
first five categories by cumulative assets under management (AUM), using tiers at
the top 70\% and 90\% of each type's global total; for VC, CVC, and angel
investors by rank cutoffs; and for investment banks by fee-revenue league-table
position.
The initial entity table lists \emph{1{,}076 institutions}; canonicalization yields
989 core entities, 201 with an SEC Central Index Key (CIK). Each record includes tier, investor type,
and an AUM estimate when publicly available. A later identity-resolution pass recovered
926 additional interface-derived entities, bringing the prepared table to
\textbf{1{,}915 investors}.

\subsection{Portfolio and Transaction Data}
\label{sec:p2d-edges}

\textbf{Portfolio edges.} For the pilot's 20 top US VCs and the scale-up stage's
broader T1+T2 VC and CVC universe, portfolio pages are scraped using per-firm CSS selector
configurations authored by an LLM-assisted generator, with a Playwright fallback for
JavaScript-rendered sites. The pilot extraction produces 2{,}281 unique $(\text{VC},
\text{company})$ edges across 17 firms;
the scale-up stage produces \emph{329{,}749 edges} across nine investor types, with per-type
coverage of 80--100\% except sovereign wealth (50\%, a confirmed disclosure
ceiling).

\textbf{Funding-event union.} Four independent pipelines fill complementary
coverage gaps. First, 592{,}032 SEC Form D filings over 47 quarters
(2015Q1--2026Q3) provide the private-placement skeleton. Special-purpose vehicle (SPV) date clustering
addresses the \emph{Rule 4(a)(2) blind spot} for exempt mega-rounds---including
those of Anthropic and OpenAI, which filed no Form D---and infers 160 round-close
windows for 69 companies. Second, news extraction (Tavily + LLM) recovers 131
high-profile rounds, including Anthropic's Series C--H history. Third, S-1 prospectus
extraction recovers 89 pre-IPO rounds for 17 of 20 portfolio companies that
completed IPOs. A
fourth pipeline extracts 492 post-IPO ownership rows from 13D/13G filings. The union is
deduplicated at the $(company, round, date)$ key and exploded to participant rows
with grain $(\text{matched investor}, \text{company}, \text{round}, \text{date})$.
Of the participant rows, 49.9\% fuzzy-match to a canonical investor; the
remainder refer to hedge funds, sovereigns, and unnamed participants outside the
entity-table scope.

\subsection{Company Canonicalization and Industry Labels}
\label{sec:p2d-companies}

Portfolio and 13F records describe the same company under many spellings
(\texttt{AMAZON COM INC}, \texttt{Amazon.com Inc}). Deduplication is driven by an
explicit, audited rule file (262 aliases) rather than fuzzy normalization; the
three retained slug collisions represent distinct securities. The result is
\textbf{43{,}474 canonical companies}\footnote{Three additional entity-building
records were excluded because both the MIT-licensed OpenSporks Crunchbase free
database snapshot downloaded from Hugging Face
\texttt{short\_description} and URL fields were empty.} with globally unique IDs and edge-conservation
checks (no transaction dropped or double-counted). Industry labels come from a
Yahoo Finance resolver (50.6\%) and category matching against that
OpenSporks snapshot (26.6\%),
merged so that 61.9\% of companies carry at least one label. The resolver validates
identity using name tokens, ranks candidates deterministically, batches lookups of CUSIP security identifiers
under rate limits, and versions cache entries for repairability and audit.

\subsection{Event-Chain Construction}
\label{sec:p2d-chains}

Construction normalizes the three event families defined in
Section~\ref{sec:p2d-structure}, resolves entity identities, and groups
participant-grain funding rows into logical outcome nodes.
Timeline resolution treats each boundary event as a
left-closed/right-open segment boundary, assigning every dated intermediate event
to exactly one chain. Raw collection, semantic review, identity patching, and
chain materialization remain separately versioned so an application can report
both row-grain provenance and graph-grain structure.

\subsection{Chip-Company Implementation}
\label{sec:p2-chip-construction}

The application in this paper begins with a 2{,}547-company chip population.
Restricting it to SUCCESS and FAILURE under the shared ontology in
Section~\ref{sec:shared-outcomes} yields a 1{,}332-company construction scope.
Collection produces 3{,}897 chain-boundary participant rows, 3{,}311 exposure
events, and 2{,}951 interface events. A later identity-resolution view contains
3{,}224 interface rows; the investor-behavior analysis uses that view rather than
the 2{,}951-row chain-construction baseline.

\begin{figure*}[t]
\centering
\includegraphics[width=0.92\textwidth]{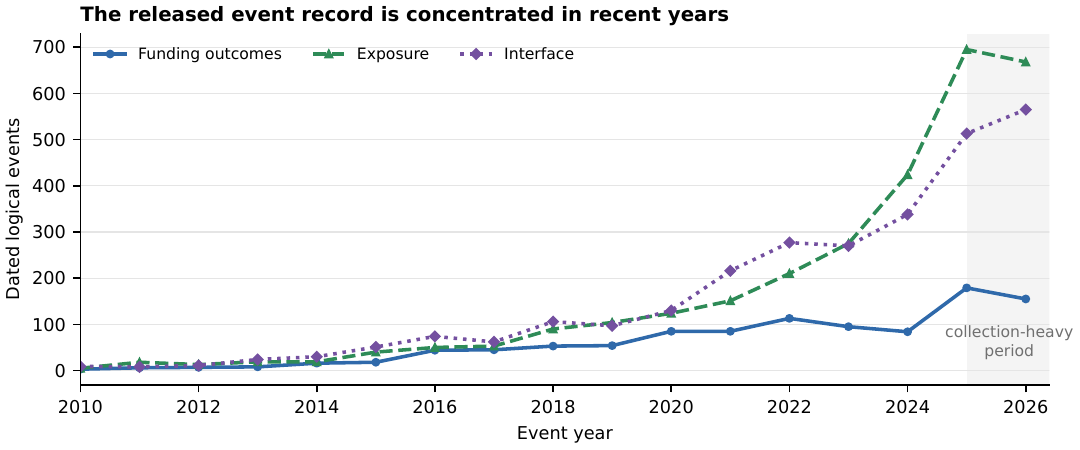}
\Description{Annual line chart of dated funding-outcome, exposure, and interface events in the released chip-company tables from 2010 through 2026.}
\caption{Temporal coverage of the released chip-company event record. Funding
participant rows are collapsed to logical company--date--type--round boundary
nodes; exposure and interface events are deduplicated by event ID. The plot
describes evidence availability rather than underlying real-world event incidence,
and the shaded 2025--2026 interval is the collection-heavy period.}
\label{fig:event-temporal-coverage}
\end{figure*}
\FloatBarrier

The annual distribution in Figure~\ref{fig:event-temporal-coverage} reveals when
the public record is informative. The
strong recent skew, especially in the 2025--2026 collection period, means older
company histories should not be read as equally observed timelines.

A 418-case human review reconciles raw event semantics before construction of the
cleaned baseline: \textbf{1{,}342 chains} (1{,}071 dated) \textbf{across 663 companies},
with event-type patches versioned separately. The Lightmatter
Series~A chain illustrates the row-grain distinction: 38 participant rows (one
lead and 37 co-investors) become one logical round node rather than 38 timeline
nodes. The Cerebras history similarly resolves all eleven chains from founding
through its 2026 IPO, with events assigned to exactly one segment.

These outputs support the two decision questions at different analytical levels:
Section~\ref{sec:part2-patterns} compares company-level outcomes across event-chain
patterns, while Section~\ref{sec:part2-behavior} examines the participating
investors' histories using the underlying investment and interface evidence.

\newcommand{\patternEstimateFigure}{%
\begin{figure*}[t]
\centering
\begin{adjustbox}{max width=\textwidth,center}
\begin{tikzpicture}[
  x=0.07cm, y=0.43cm,
  font=\figurebodyfont,
  every node/.style={font=\figurebodyfont},
  point/.style={draw=black!45, fill=white, circle, inner sep=1.4pt},
  pointOK/.style={fill=reportGreen, draw=reportGreen, circle, inner sep=1.4pt},
  pointAVOID/.style={fill=reportRed, draw=reportRed, circle, inner sep=1.4pt},
]
\draw[thick, black] (0,0.3) -- (0,8.0);
\node[above] at (0,8.0) {0 pp};

\draw[->, thick] (-55,0.2) -- (45,0.2) node[right] {$\Delta_P$ (pp)};
\foreach \x in {-40,-20,0,20,40}{
  \draw[black!35] (\x,0.2)--(\x,-0.05);
  \node[below] at (\x,-0.1) {\x};
}

\draw[black!45, line width=0.8pt] (-0.44,7.15) -- (17.74,7.15);
\node[point] at (8.7,7.15) {};
\node[anchor=west, black!55] at (12.5,7.40) {P1:$+8.7$\,pp INCONCLUSIVE};

\draw[reportGreen, line width=0.8pt] (20.97,6.15) -- (41.50,6.15);
\node[pointOK] at (32.8,6.15) {};
\node[anchor=west, reportGreen] at (36.5,6.40) {P2:$+32.8$\,pp RECOMMEND};

\draw[reportGreen, line width=0.8pt] (5.70,5.15) -- (24.69,5.15);
\node[pointOK] at (15.6,5.15) {};
\node[anchor=west, reportGreen] at (19.5,5.40) {P3:$+15.6$\,pp RECOMMEND};

\draw[black!45, line width=0.8pt] (-5.42,4.15) -- (36.63,4.15);
\node[point] at (20.25,4.15) {};
\node[anchor=west, black!55] at (24.0,4.40) {P4:$+20.3$\,pp INCONCLUSIVE};

\draw[black!45, line width=0.8pt] (-15.10,3.15) -- (9.84,3.15);
\node[point] at (-2.6,3.15) {};
\node[anchor=west, black!55] at (2.0,3.40) {P5:$-2.6$\,pp INCONCLUSIVE};

\draw[black!45, line width=0.8pt] (-5.57,2.15) -- (26.35,2.15);
\node[point] at (11.82,2.15) {};
\node[anchor=west, black!55] at (15.5,2.40) {P7:$+11.8$\,pp INCONCLUSIVE};

\draw[reportRed, line width=0.8pt] (-50.67,1.15) -- (-9.45,1.15);
\node[pointAVOID] at (-37.9,1.15) {};
\node[anchor=west, reportRed] at (-33.5,1.40) {P8:$-37.9$\,pp AVOID};
\end{tikzpicture}
\end{adjustbox}
\Description{Forest-style plot of unadjusted SUCCESS-share differences for seven interpretable event-chain patterns.}
\caption{First-batch pattern estimate plot: unadjusted difference in observed
SUCCESS shares, $\Delta_P=\hat p_1-\hat p_0$, in percentage points. Filled
green points carry the \texttt{RECOMMEND} registry label under the unadjusted threshold, filled
red the \texttt{AVOID} label, and grey hollow points carry \texttt{INCONCLUSIVE}.
Horizontal lines are Newcombe 95\% confidence intervals.
\texttt{P2} ($+32.8$\,pp), \texttt{P3}
($+15.6$\,pp) and \texttt{P8} ($-37.9$\,pp) receive directional registry labels;
\texttt{P1}, \texttt{P4}, \texttt{P5}, and \texttt{P7} carry \texttt{INCONCLUSIVE};
\texttt{P6} is not evaluable in the current vocabulary.}
\label{fig:pattern-ate}
\end{figure*}
}

\section{Event-Chain Pattern Analysis}
\label{sec:part2-patterns}

This section addresses the first founder question: which external operating
actions should be prioritized when working toward the next financing milestone?
It examines associations between completed event-chain patterns and the shared
company-level outcome labels, rather than predicting the next round itself.
The chip-company findings distinguish sustained operating progress from
financing without progress, and interpret investor participation and governance
changes in context.

\subsection{Method Design}
\label{sec:p2-pattern-method}

A \emph{pattern} is a reproducible classification rule over the temporally
ordered events constructed in Section~\ref{sec:p2d-chains}, including a terminal
outcome node only where the definition explicitly calls for it. The method
estimates each pattern's association with the company-level outcome label; it is retrospective
pattern discovery, not a simulation of information available at an earlier
decision time.
Classical sequential-pattern mining discovers frequent ordered subsequences from
sequence databases~\cite{pei2001prefixspan}; our bounded grammar instead freezes
business-interpretable event predicates before outcome-label evaluation.

Definitions live in an external, schema-validated rule configuration rather than in
Python. The engine rejects unknown fields, operators, and pattern references, and
each run records the resolved configuration's SHA-256, schema version, and
pattern-set ID. Candidate patterns are generated from a frozen observed-domain
grammar without seeing outcome labels. Support pruning and duplicate-company-mask
deduplication reduce redundant candidates before statistical evaluation, and
accepted definitions are consolidated into a canonical registry.

Within the evaluable company cohort, $P=1$ if at least one dated, evaluable chain
matches pattern $P$, and $P=0$ otherwise. Each company is counted once per pattern,
regardless of how many chains match.

For each operating-action or event-chain pattern $P$, the method reports the
unadjusted difference in observed SUCCESS shares,
$\Delta_P=\hat p(Y=1\mid P=1)-\hat p(Y=1\mid P=0)$, with a pooled
two-proportion z-test for $H_0:\Delta_P=0$ and the two one-sided tests (TOST)
procedure with a $\pm5$ percentage-point equivalence margin. If the two-sided
test yields $p<0.05$, a positive difference receives \texttt{RECOMMEND} and a
negative difference \texttt{AVOID}. Otherwise, TOST equivalence at the 0.05 level
yields \texttt{INDIFFERENT}; if neither test passes, the label is
\texttt{INCONCLUSIVE}. These labels apply to an observed action or pattern, never
to a company, and denote exploratory associations rather than causal treatment
recommendations, forecasts, or investment advice.
The pooled score test remains the frozen registry rule. For the interpretable
first-batch patterns, we additionally report Newcombe 95\% confidence intervals
for the absolute risk difference and two-sided $p$-values from Fisher's exact test as a
small-cell sensitivity analysis~\cite{newcombe1998}; these diagnostics do not
relabel the registry.

\subsection{Chip-Company Evaluation Scope}
\label{sec:p2p-design}

\textbf{Evaluation population.} The analysis covers the 501 chip-universe companies
with at least one dated, evaluable chain: 269 SUCCESS and 232 FAILURE. This is a
differentially observable subset: 269 of 390 SUCCESS companies (69\%) but only
232 of 942 FAILURE companies (25\%) in the underlying labeled scope are
evaluable.

\begin{figure*}[t]
\centering
\includegraphics[width=0.71\textwidth]{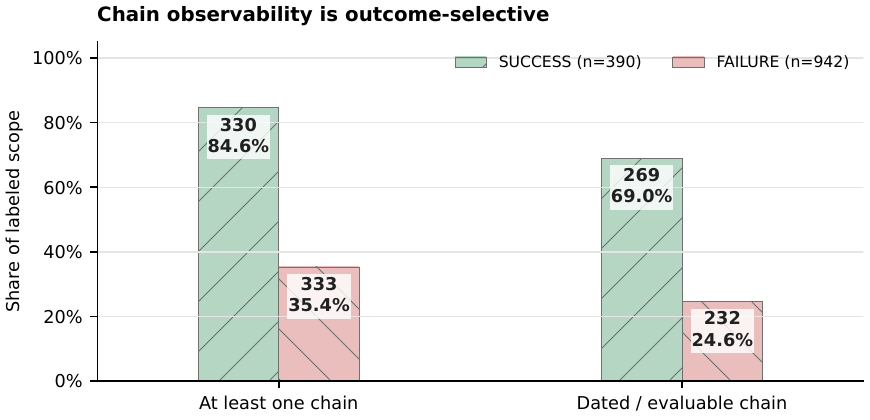}
\Description{Grouped bars comparing chain presence and dated-chain evaluability among SUCCESS and FAILURE companies in the labeled chip scope.}
\caption{Differential public-source chain observability by company outcome. Of
390 SUCCESS companies, 330
have at least one chain and 269 have a dated and evaluable chain; the corresponding
counts among 942 FAILURE companies are 333 and 232. Consequently, downstream
pattern estimates describe a selected subset rather than the full labeled scope.}
\label{fig:chain-observability}
\end{figure*}
\FloatBarrier

Figure~\ref{fig:chain-observability} quantifies this coverage difference.
Successful companies generally leave more financing announcements, product and
customer milestones, regulatory filings, and media coverage, whereas failed
companies may cease operating without a comparably visible public trail. The
gap is therefore a limitation of reconstruction from public sources: it does
not indicate outcome-based sampling by the researchers, but it limits how far
the observable subset can represent the entire labeled scope.

\subsection{Operating Progress, Financing, and Investor Involvement}
\label{sec:p2p-p1p8}

The findings below combine the eight initial hypotheses (P1--P8) with the
label-blind enumeration results. All reported differences are unadjusted
percentage-point differences in observed SUCCESS shares. The initial hypotheses
provide interpretable reference contrasts (Table~\ref{tab:first-batch-patterns});
the enumerated patterns test more specific combinations.

\textbf{Operating progress and sustained milestones.} Product, customer, or
supply-chain action (P3) is associated with a higher SUCCESS share
($+15.6$\,pp). The retained enumeration results also favor a product launch
($+19.9$\,pp), at least two intermediate events ($+17.4$\,pp), and at least two
exposure events ($+16.6$\,pp). These patterns carry \texttt{RECOMMEND} labels.
Together, they support attention to operating progress and milestone
follow-through, rather than treating one announcement as a complete strategy.

\textbf{Financing across long event spans and gaps.} A span of at least three
years from the first to the last intermediate event, combined with funding
participation, carries an \texttt{AVOID} label ($-20.1$\,pp). A distinct pattern
combines a longest gap of at least three years with at most one general-funding
event ($-21.5$\,pp), also labeled \texttt{AVOID}. The first measures the span of
the recorded history, not inactivity throughout that period; the second measures
an interval without recorded events. Financing participation alone does not
establish continuing operating progress.

\textbf{Investor involvement in context.}
Strategic investment ($+22.9$\,pp) and the presence of an interface event
($+12.9$\,pp) are the other two retained \texttt{RECOMMEND} patterns.
Broad terminal-round participation also co-occurs with success even without
recorded intermediate actions: P2 measures at least five investors at the
terminal boundary and yields $+32.8$\,pp. Thus, visible actions alone do not
exhaust the information in investor participation. Conversely, a major-holder
change with sparse activity carries an \texttt{AVOID} label ($-29.4$\,pp).
P8 likewise combines holder change or litigation with missing operating
follow-through ($-37.9$\,pp); its exact exclusions appear below the table.
The warning concerns the surrounding event sequence, not ownership or governance
change in isolation. P1, P4, P5, and P7 remain inconclusive; P6 is not evaluable
in the current vocabulary.

Section~\ref{sec:p2p-batches} distinguishes these frozen registry results from
their multiplicity and coverage/time sensitivity checks.

\ifdefined\TenPtClose
\begin{figure*}[t]
\centering
\begin{adjustbox}{max width=\textwidth,center}
\begin{tikzpicture}[
  x=0.07cm, y=0.43cm,
  font=\figurebodyfont,
  every node/.style={font=\figurebodyfont},
  point/.style={draw=black!45, fill=white, circle, inner sep=1.4pt},
  pointOK/.style={fill=reportGreen, draw=reportGreen, circle, inner sep=1.4pt},
  pointAVOID/.style={fill=reportRed, draw=reportRed, circle, inner sep=1.4pt},
]
\draw[thick, black] (0,0.3) -- (0,8.0);
\node[above] at (0,8.0) {0 pp};

\draw[->, thick] (-55,0.2) -- (45,0.2) node[right] {$\Delta_P$ (pp)};
\foreach \x in {-40,-20,0,20,40}{
  \draw[black!35] (\x,0.2)--(\x,-0.05);
  \node[below] at (\x,-0.1) {\x};
}

\draw[black!45, line width=0.8pt] (-0.44,7.15) -- (17.74,7.15);
\node[point] at (8.7,7.15) {};
\node[anchor=west, black!55] at (12.5,7.40) {P1:$+8.7$\,pp INCONCLUSIVE};

\draw[reportGreen, line width=0.8pt] (20.97,6.15) -- (41.50,6.15);
\node[pointOK] at (32.8,6.15) {};
\node[anchor=west, reportGreen] at (36.5,6.40) {P2:$+32.8$\,pp RECOMMEND};

\draw[reportGreen, line width=0.8pt] (5.70,5.15) -- (24.69,5.15);
\node[pointOK] at (15.6,5.15) {};
\node[anchor=west, reportGreen] at (19.5,5.40) {P3:$+15.6$\,pp RECOMMEND};

\draw[black!45, line width=0.8pt] (-5.42,4.15) -- (36.63,4.15);
\node[point] at (20.25,4.15) {};
\node[anchor=west, black!55] at (24.0,4.40) {P4:$+20.3$\,pp INCONCLUSIVE};

\draw[black!45, line width=0.8pt] (-15.10,3.15) -- (9.84,3.15);
\node[point] at (-2.6,3.15) {};
\node[anchor=west, black!55] at (2.0,3.40) {P5:$-2.6$\,pp INCONCLUSIVE};

\draw[black!45, line width=0.8pt] (-5.57,2.15) -- (26.35,2.15);
\node[point] at (11.82,2.15) {};
\node[anchor=west, black!55] at (15.5,2.40) {P7:$+11.8$\,pp INCONCLUSIVE};

\draw[reportRed, line width=0.8pt] (-50.67,1.15) -- (-9.45,1.15);
\node[pointAVOID] at (-37.9,1.15) {};
\node[anchor=west, reportRed] at (-33.5,1.40) {P8:$-37.9$\,pp AVOID};
\end{tikzpicture}
\end{adjustbox}
\Description{Forest-style plot of unadjusted SUCCESS-share differences for seven interpretable event-chain patterns.}
\caption{First-batch pattern estimate plot: unadjusted difference in observed
SUCCESS shares, $\Delta_P=\hat p_1-\hat p_0$, in percentage points. Filled
green points carry the \texttt{RECOMMEND} registry label under the unadjusted threshold, filled
red the \texttt{AVOID} label, and grey hollow points carry \texttt{INCONCLUSIVE}.
Horizontal lines are Newcombe 95\% confidence intervals.
\texttt{P2} ($+32.8$\,pp), \texttt{P3}
($+15.6$\,pp) and \texttt{P8} ($-37.9$\,pp) receive directional registry labels;
\texttt{P1}, \texttt{P4}, \texttt{P5}, and \texttt{P7} carry \texttt{INCONCLUSIVE};
\texttt{P6} is not evaluable in the current vocabulary.}
\label{fig:pattern-ate}
\end{figure*}

\fi

\begin{center}
\begin{minipage}{\textwidth}
\centering
\captionof{table}{First-batch event-chain patterns and registry labels.}
\label{tab:first-batch-patterns}
\scriptsize
\renewcommand{\arraystretch}{1.15}
\setlength{\tabcolsep}{2.2pt}
\begin{adjustbox}{max width=\textwidth,center}
\begin{tabular}{lccccc}
\toprule
Pattern & \shortstack{SUCCESS/total\\(present; absent)} & \shortstack{$\Delta_P$ [95\% CI]\\(percentage points)} & Score $p$ & Fisher $p$ & Registry label \\
\midrule
P1: no recorded action & 186/328; 83/173 & $+8.73$ [$-0.44,17.74$] & 0.0624 & 0.0733 & \texttt{INCONCLUSIVE} \\
P2: $\ge$5 terminal-round investors, no action & 55/67; 214/434 & $+32.78$ [$20.97,41.50$] & $5.49\times10^{-7}$ & $3.08\times10^{-7}$ & \textbf{RECOMMEND} \\
P3: product, customer, or supply action & 86/132; 183/369 & $+15.56$ [$5.70,24.69$] & 0.00209 & 0.00227 & \textbf{RECOMMEND} \\
P4: regulatory or export action & 11/15; 258/486 & $+20.25$ [$-5.42,36.63$] & 0.1214 & 0.187 & \texttt{INCONCLUSIVE} \\
P5: distress or restructuring & 35/68; 234/433 & $-2.57$ [$-15.10,9.84$] & 0.6926 & 0.697 & \texttt{INCONCLUSIVE} \\
P6: acquisition or debt refinancing & \multicolumn{4}{c}{not evaluable in current vocabulary} & \texttt{NOT\_EVALUABLE} \\
P7: board change without P3--P5 & 22/34; 247/467 & $+11.82$ [$-5.57,26.35$] & 0.1822 & 0.214 & \texttt{INCONCLUSIVE} \\
P8: holder change or litigation, with exclusions & 2/12; 267/489 & $-37.93$ [$-50.67,-9.45$] & 0.00922 & 0.0154 & \textbf{AVOID} \\
\bottomrule
\end{tabular}
\end{adjustbox}
\par
\noindent P8 requires a major-holder change or litigation, with no product-launch,
customer-change, supply-chain, or interface-exit event in the same chain.
\end{minipage}
\end{center}

Figure~\ref{fig:pattern-ate} visualizes the seven evaluable first-batch estimates.
\ifdefined\TenPtClose\else
\begin{figure*}[t]
\centering
\begin{adjustbox}{max width=\textwidth,center}
\begin{tikzpicture}[
  x=0.07cm, y=0.43cm,
  font=\figurebodyfont,
  every node/.style={font=\figurebodyfont},
  point/.style={draw=black!45, fill=white, circle, inner sep=1.4pt},
  pointOK/.style={fill=reportGreen, draw=reportGreen, circle, inner sep=1.4pt},
  pointAVOID/.style={fill=reportRed, draw=reportRed, circle, inner sep=1.4pt},
]
\draw[thick, black] (0,0.3) -- (0,8.0);
\node[above] at (0,8.0) {0 pp};

\draw[->, thick] (-55,0.2) -- (45,0.2) node[right] {$\Delta_P$ (pp)};
\foreach \x in {-40,-20,0,20,40}{
  \draw[black!35] (\x,0.2)--(\x,-0.05);
  \node[below] at (\x,-0.1) {\x};
}

\draw[black!45, line width=0.8pt] (-0.44,7.15) -- (17.74,7.15);
\node[point] at (8.7,7.15) {};
\node[anchor=west, black!55] at (12.5,7.40) {P1:$+8.7$\,pp INCONCLUSIVE};

\draw[reportGreen, line width=0.8pt] (20.97,6.15) -- (41.50,6.15);
\node[pointOK] at (32.8,6.15) {};
\node[anchor=west, reportGreen] at (36.5,6.40) {P2:$+32.8$\,pp RECOMMEND};

\draw[reportGreen, line width=0.8pt] (5.70,5.15) -- (24.69,5.15);
\node[pointOK] at (15.6,5.15) {};
\node[anchor=west, reportGreen] at (19.5,5.40) {P3:$+15.6$\,pp RECOMMEND};

\draw[black!45, line width=0.8pt] (-5.42,4.15) -- (36.63,4.15);
\node[point] at (20.25,4.15) {};
\node[anchor=west, black!55] at (24.0,4.40) {P4:$+20.3$\,pp INCONCLUSIVE};

\draw[black!45, line width=0.8pt] (-15.10,3.15) -- (9.84,3.15);
\node[point] at (-2.6,3.15) {};
\node[anchor=west, black!55] at (2.0,3.40) {P5:$-2.6$\,pp INCONCLUSIVE};

\draw[black!45, line width=0.8pt] (-5.57,2.15) -- (26.35,2.15);
\node[point] at (11.82,2.15) {};
\node[anchor=west, black!55] at (15.5,2.40) {P7:$+11.8$\,pp INCONCLUSIVE};

\draw[reportRed, line width=0.8pt] (-50.67,1.15) -- (-9.45,1.15);
\node[pointAVOID] at (-37.9,1.15) {};
\node[anchor=west, reportRed] at (-33.5,1.40) {P8:$-37.9$\,pp AVOID};
\end{tikzpicture}
\end{adjustbox}
\Description{Forest-style plot of unadjusted SUCCESS-share differences for seven interpretable event-chain patterns.}
\caption{First-batch pattern estimate plot: unadjusted difference in observed
SUCCESS shares, $\Delta_P=\hat p_1-\hat p_0$, in percentage points. Filled
green points carry the \texttt{RECOMMEND} registry label under the unadjusted threshold, filled
red the \texttt{AVOID} label, and grey hollow points carry \texttt{INCONCLUSIVE}.
Horizontal lines are Newcombe 95\% confidence intervals.
\texttt{P2} ($+32.8$\,pp), \texttt{P3}
($+15.6$\,pp) and \texttt{P8} ($-37.9$\,pp) receive directional registry labels;
\texttt{P1}, \texttt{P4}, \texttt{P5}, and \texttt{P7} carry \texttt{INCONCLUSIVE};
\texttt{P6} is not evaluable in the current vocabulary.}
\label{fig:pattern-ate}
\end{figure*}
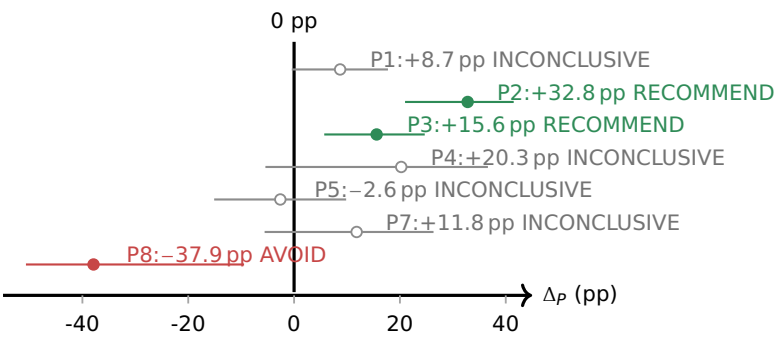

\fi
\FloatBarrier

\subsection{Registry Scope and Robustness}
\label{sec:p2p-batches}

\textbf{Registry scope.} The three batches yield \textbf{23{,}308 patterns}:
9{,}460 RECOMMEND, 75 AVOID, and 13{,}773 INCONCLUSIVE; one unobservable definition
is excluded from that total. Selection for the interpretable shortlist requires
coherent business meaning, not significance alone. The five retained batch-2
findings all have $p<0.005$. Registry labels use unadjusted tests: label-blind
generation does not eliminate multiplicity, and overlapping patterns must not be
read as independent findings.

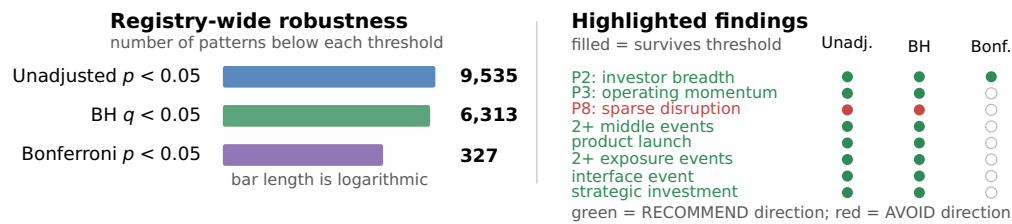
\begin{figure*}[t]
\centering
\begin{adjustbox}{max width=0.97\textwidth,center}
\begin{tikzpicture}[report figure, x=1cm, y=0.46cm]
  \node[report title,anchor=west] at (0,1.0) {Registry-wide robustness};
  \node[report note,anchor=west] at (0,0.3) {number of patterns below each threshold};
  \foreach \y/\label/\count/\width/\clr in {
    -0.8/{Unadjusted $p<0.05$}/{9,535}/3.25/reportBlue,
    -2.1/{BH $q<0.05$}/{6,313}/3.17/reportGreen,
    -3.4/{Bonferroni $p<0.05$}/{327}/2.45/reportPurple}{
      \node[anchor=east] at (1.65,\y) {\label};
      \path[fill=\clr!78,rounded corners=1pt] (1.85,\y-0.36) rectangle +(\width,0.72);
      \node[anchor=west,font=\figurebodyfont\bfseries] at (5.35,\y) {\count};
  }
  \node[report note,anchor=west] at (1.85,-4.25) {bar length is logarithmic};

  \draw[black!18] (6.65,1.05) -- (6.65,-4.35);

  \node[report title,anchor=west] at (7.05,1.0) {Highlighted findings};
  \node[report note,anchor=west] at (7.05,0.3) {filled = survives threshold};
  \foreach \x/\head in {11.4/{Unadj.},12.5/{BH},13.6/{Bonf.}}{
    \node[font=\figurenotefont,anchor=south] at (\x,-0.25) {\head};
  }
  \foreach \y/\label/\clr/\bonf in {
    -0.8/{P2: investor breadth}/reportGreen/1,
    -1.35/{P3: operating momentum}/reportGreen/0,
    -1.9/{P8: sparse disruption}/reportRed/0,
    -2.45/{2+ middle events}/reportGreen/0,
    -3.0/{product launch}/reportGreen/0,
    -3.55/{2+ exposure events}/reportGreen/0,
    -4.1/{interface event}/reportGreen/0,
    -4.65/{strategic investment}/reportGreen/0}{
      \node[anchor=west,font=\figurenotefont,text=\clr] at (7.05,\y) {\label};
      \fill[\clr] (11.4,\y) circle (2.4pt);
      \fill[\clr] (12.5,\y) circle (2.4pt);
      \ifnum\bonf=1 \fill[\clr] (13.6,\y) circle (2.4pt);
      \else \draw[black!28,line width=0.5pt] (13.6,\y) circle (2.4pt); \fi
  }
  \node[report note,anchor=west] at (7.05,-5.35)
    {green = RECOMMEND direction; red = AVOID direction};
\end{tikzpicture}
\end{adjustbox}
\Description{Registry-wide counts surviving unadjusted, Benjamini--Hochberg, and Bonferroni thresholds, followed by a matrix showing that P2 survives both corrections and seven other highlighted findings survive Benjamini--Hochberg.}
\caption{Multiplicity robustness of the frozen pattern registry. Of 9{,}535
nominally significant patterns, 6{,}313 remain under Benjamini--Hochberg control and
327 under Bonferroni correction. P2 survives both; P3, P8, and all five retained
batch-2 findings survive Benjamini--Hochberg. Directional colors retain the
registry's action interpretation and are separate from correction survival.}
\label{fig:multiplicity-robustness}
\end{figure*}

\textbf{Multiplicity sensitivity.} Post hoc Benjamini--Hochberg (BH)
correction~\cite{benjamini1995fdr} at $q<0.05$ retains P2, P3, P8, and all five
batch-2 highlights; P2 also survives Bonferroni correction. P1, P4, P5, and P7
remain non-significant. Figure~\ref{fig:multiplicity-robustness} reports the
registry-wide survival counts. These checks provide robustness evidence without
changing the frozen registry labels.

\textbf{Coverage/time sensitivity.} On the released identity-patched
snapshot,\footnote{The frozen registry predates the Interface-identity patch.
Seven of eight highlighted raw $2\times2$ tables reproduce exactly. For
``at least two middle events,'' five company assignments change: frozen
144/84 versus 125/148 becomes released 147/86 versus 122/146
(SUCCESS/FAILURE among pattern-positive versus pattern-negative companies).
The direction and unadjusted significance are unchanged. The reproduction
script uses the released snapshot as canonical.}
the eight highlights are stratified by dated-chain count (1, 2, or 3+) and first
dated outcome year ($\leq2015$, 2016--2020, or $\geq2021$).
Cochran--Mantel--Haenszel common odds ratios retain all eight directions, but
only P2 (OR 2.19, $p=0.0248$) and P8 (OR 0.16, $p=0.00338$) remain nominally
significant. After BH correction across these eight checks, only P8 remains
significant ($q=0.027$; P2: $q=0.099$). P3 and the five batch-2 highlights lose
significance under the coverage/time adjustment. P8 is therefore the most robust
highlight in this check; positive operating-momentum findings remain
hypothesis-generating. This is not causal adjustment: founding year and
unobserved private activity are unavailable.

\FloatBarrier

\subsection{Decision-Action Interpretation}
\label{sec:p2p-interpretation}

For a founder planning the next financing milestone, the chip-company evidence favors
\emph{building and documenting sustained operating progress}: product delivery,
customer development, and supply-chain milestones deserve attention alongside
fundraising. Continued financing is not a substitute for that progress, and
ownership or governance events should be investigated in the context of the
surrounding operating record. Investor participation should likewise be assessed
for its substantive role rather than by visible interaction count alone.
The \texttt{RECOMMEND}/\texttt{AVOID} action-pattern labels help prioritize
evidence gathering and operating review, not automate decisions.
Section~\ref{sec:part2-behavior} addresses the complementary question:
whom to approach for capital and strategic support.

\newcommand{\confirmedBehaviorFigure}{%
\begin{figure*}[t]
\centering
\includegraphics[width=0.90\textwidth]{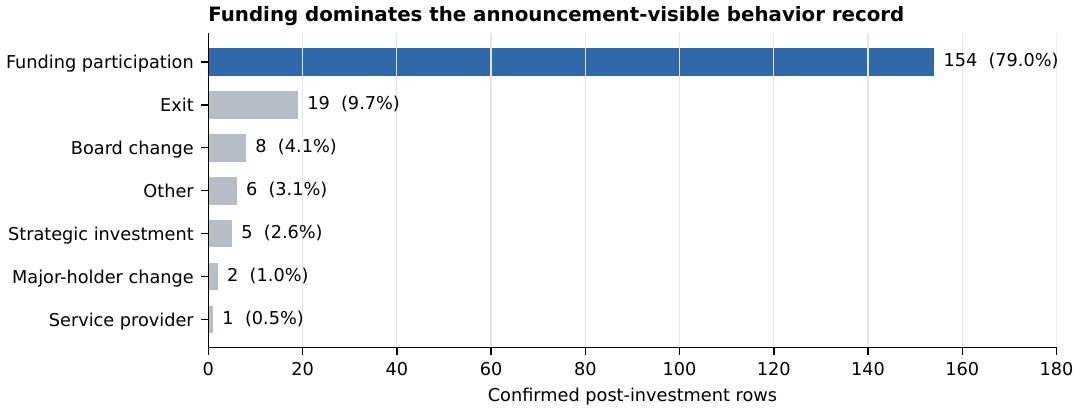}
\Description{Horizontal bars showing the count and share of all 195 confirmed post-investment rows in each behavior category.}
\caption{Composition of the 195 confirmed post-investment rows. Funding
participation accounts for 79.0\%; the non-funding tail is led by 19 exit rows and
is otherwise sparse. The figure reports announcement-visible evidence, not the
true incidence of private investor actions.}
\label{fig:confirmed-behavior-composition}
\end{figure*}
}

\section{Investor Post-Investment Behavior Analysis}
\label{sec:part2-behavior}

This section addresses the second founder question: which investors or
investment institutions should be approached for the company's intended
trajectory? It examines investor histories within the same event-chain evidence
framework, focusing on continued financing, strategic participation, acquisition,
and governance. The analysis moves from confirmed post-investment actions to
investor-type profiles and then individual histories, so founders can distinguish
capital continuity from strategic or ownership involvement.

\subsection{Method Design}
\label{sec:p2-behavior-method}

Investment evidence is joined to interface events on
$(\text{investor},\text{company})$, retaining an event-level temporal audit trail.
This computation uses the underlying records, not the materialized chain table
or the pattern labels from Section~\ref{sec:part2-patterns}.
Dates are represented as intervals so partially specified dates can be
handled conservatively. An action is \emph{confirmed post-investment} only when
the upper bound of the investment interval is strictly earlier than the lower
bound of the interface-event interval. Other records are classified as
same or overlapping date, timing unknown, no investment evidence, or investment
evidence later than the interface event.

Date-keyed actions are deduplicated into episodes and aggregated independently
at event, investor--company pair, investor, and investor-type levels. Composition
among confirmed actions is reported separately from profile-level observation
coverage.

\ifdefined\TenPtClose
\confirmedBehaviorFigure
\fi

\subsection{Chip-Company Results: Confirmed Actions}
\label{sec:p2b-confirmed}

Of 3{,}224 interface rows, 1{,}543 have a resolved investor. Applying the strict
interval rule, the temporal join admits
\textbf{195 confirmed rows} --- 194 date-keyed episodes, 134 investor--company pairs,
and 119 investors.\footnote{The other resolved rows have same or overlapping
dates (817), unknown timing (121), no investment evidence (402), or investment
evidence only after the interface event (8).}

Within the announcement-visible confirmed subset, the dominant pattern is
\textbf{capital continuity, not operational control}:
funding participation accounts for 154/195 confirmed rows (79.0\%), and at the
pair level 101/134 pairs (75.4\%) show funding participation and nothing else. Only
eight pairs combine follow-on funding with an exit.
Figure~\ref{fig:confirmed-behavior-composition} shows the composition of these
confirmed rows.

\ifdefined\TenPtClose\else
\confirmedBehaviorFigure
\fi
\FloatBarrier

The remaining 41 rows span six behavior types; exit is the only non-funding
category with more than ten confirmed observations.

\subsection{Investor-Type Profiles}
\label{sec:p2b-types}

For descriptive comparison, we display investor types with at least five
confirmed non-funding rows. This is a support rule, not a significance test or
a claim of representative coverage. Three types meet it:

\begin{itemize}
  \item \textbf{Institutional VC --- financing-led capital continuity.} Of 552
        profiles, 64 (11.6\%) have confirmed behavior; 110 confirmed rows are 85.5\%
        funding events. 52/64 confirmed institutional VCs are observed through
        funding only.
  \item \textbf{CVC --- capital support with strategic and exit options.}
        Of 211 profiles, 29 (13.7\%) have confirmed behavior. Funding is still the
        largest component (67.3\%), but 11/29 confirmed CVCs show at least one
        non-funding behavior --- exits, strategic investments, board changes --- a
        broader observed channel than institutional VC.
  \item \textbf{PE --- realization and governance.} Of 60 profiles, 9 (15.0\%) have
        confirmed behavior; non-funding behavior is the majority (53.8\%: five exits,
        two board changes). Only 3/9 confirmed PE investors are funding-only;
        the nine-investor sample makes the conclusion provisional.
\end{itemize}

Other types have insufficient confirmed non-funding evidence for stable
comparisons; hedge funds, pensions, and accelerators have no confirmed
post-investment actions in the observation window.

\begin{figure*}[t]
\centering
\includegraphics[width=0.94\textwidth]{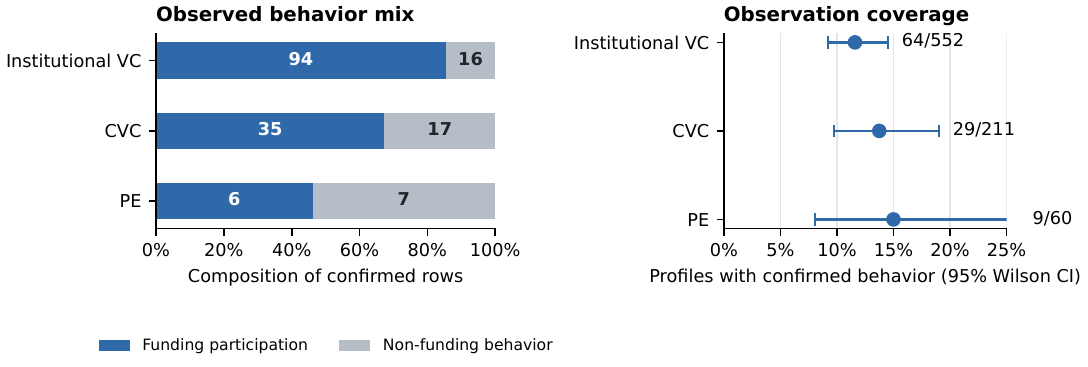}
\Description{Two-panel investor-type comparison: stacked funding versus non-funding composition of confirmed rows, and confirmed-profile coverage with Wilson intervals.}
\caption{Observed behavior mix and observation coverage for the three investor
types with enough confirmed non-funding evidence to compare. Counts inside the
stacked bars are confirmed rows. The right panel reports the share of profiles
with any confirmed behavior and Wilson 95\% intervals, emphasizing the limited
support behind the type-level composition.}
\label{fig:investor-type-behavior}
\end{figure*}
\FloatBarrier

Figure~\ref{fig:investor-type-behavior} separates behavior composition among
confirmed rows from profile-level observation coverage, with Wilson intervals
showing the uncertainty around coverage.

\subsection{Individual Histories: Seven Support-Gated Investors}
\label{sec:p2b-gate}

Individual histories distinguish strategies that a type label alone can obscure.
This comparison uses the full interface histories of all 1{,}025 resolved
investors, not only the 119 with confirmed post-investment actions; its episodes
need not satisfy the strict temporal rule and are not part of the 195-row
composition. A uniform support gate requires at least three deduplicated core
episodes and at least three interface companies, without a name filter or
significance test. Core episodes comprise exits, strategic investments, board
changes, major-holder changes, secondary transactions, and activist actions,
deduplicated by investor, company, event type, and date. Seven investors covering
37 chip companies qualify, with 28 core episodes: 14 exits (50\%), 6 strategic
investments (21.4\%), and 8 governance or ownership episodes (28.6\%).

\begin{figure*}[t]
\centering
\includegraphics[width=0.66\textwidth]{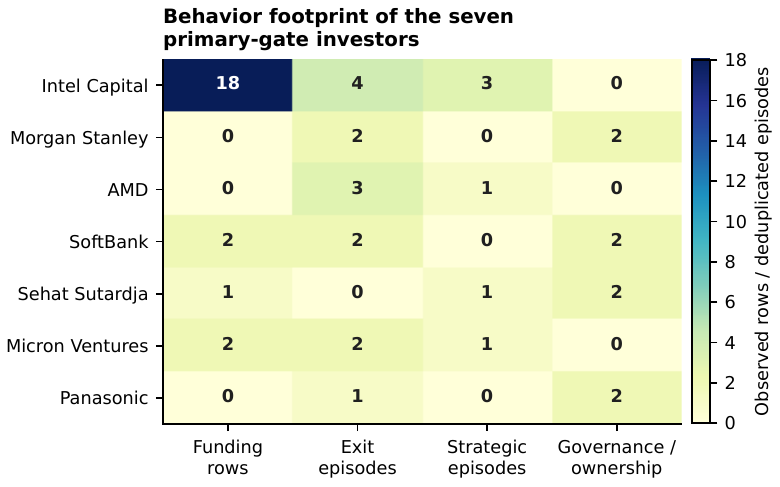}
\Description{Heat map of funding rows and exit, strategic, and governance or ownership episodes for the seven primary-gate investors.}
\caption{Behavior footprint of the seven primary-gate investors. Funding is
reported as source rows; the other columns are deduplicated core episodes. The
heat map exposes the heterogeneous scale and composition that underlie the three
qualitative behavior groupings.}
\label{fig:gate-investor-heatmap}
\end{figure*}
\FloatBarrier

Figure~\ref{fig:gate-investor-heatmap} reveals Intel Capital's much larger funding
footprint and the sparse support for several other profiles before the investors
are grouped into narrative patterns.

Their full event sets form three descriptive patterns:

\begin{itemize}
  \item \textbf{Technology investment and capability acquisition} --- Intel Capital,
        AMD, Micron Ventures, and Sehat Sutardja. The three corporates combine
        strategic investment with acquisition of chip or compute companies (Intel
        invests broadly then acquires selectively; AMD favors direct team or company
        acquisition; Micron pairs a strategic investment in Mythic with an
        acquisition); Sutardja is the non-acquisition subtype (strategic investment plus
        founder or board links).
  \item \textbf{Financial ownership management and realization} --- Morgan Stanley
        and Panasonic. Morgan Stanley's trajectory is position management (IPO, stake
        increase, H-share purchase); Panasonic's is restructuring and exit (spin-out,
        IPO, sale of a 49\% stake). Both act through ownership and capital-market
        events rather than technology acquisition.
  \item \textbf{Acquisition followed by control and integration} --- SoftBank. Its
        four core episodes all concern one company (Graphcore): acquisition, then a
        post-acquisition capital injection and a management or founder change --- a
        concentrated acquisition-to-integration sequence rather than a portfolio-wide
        pattern.
\end{itemize}

\subsection{Decision Implications and Limitations}
\label{sec:p2b-limits}

For founders seeking acquisition,
the evidence tentatively favors approaching strategic corporate investors with
relevant acquisition histories; the Intel, AMD, and Micron cases illustrate the
technology-acquisition channel to examine. For founders prioritizing independent
operation, financing-led institutional VCs with fewer observable control events
are a more relevant starting point for outreach. The SoftBank case illustrates
why acquisition followed by integration differs from continued financial support.
These are conditional starting points, not investor rankings: individual histories
and their fit with the founder's objective matter more than institutional labels.

The evidence is descriptive, not causal or predictive, and public records favor
announced financing over less-visible governance activity. Missing control events
do not establish a hands-off relationship; behavior composition and type-level
differences may change as coverage improves.
Section~\ref{sec:general-limitations} consolidates the quantitative coverage,
missingness, and collection-history constraints.

\bookmarksetupnext{startatroot}
\section{General Discussion}
\label{sec:conclusion}

Taken together, Part~One shows that a frozen, evidence-backed proposal pipeline
can be evaluated against historical outcomes, while Part~Two shows that public
records can be transformed into auditable event chains for retrospective
analysis. Their common contribution is not a single model of venture success,
but a reusable evidence discipline that makes sources, time bounds, provenance,
and uncertainty explicit at two founder decision stages.

\textbf{Professional databases and the public-data alternative.} Subscription
venture databases remain valuable professional infrastructure: their curated
coverage, standardized records, and research interfaces can be appropriate for
institutions able to purchase them. Their price and access conditions, however,
may make them a poor default for individual founders and resource-constrained
researchers. The public-source toolchain is therefore a complementary access
path, not a claim that free retrieval dominates professional databases. It has
two purposes: to test whether comparable decision-support research is feasible
from public evidence while exposing the resulting coverage limits, and to
release the construction and analysis methods for methodological scrutiny,
reuse, and scholarly discussion.

\textbf{Data availability before model choice.} The choice of an LLM pipeline
over a newly trained predictive model reflects the bottleneck addressed here,
not a claim that pipelines dominate learned models. The immediate question is
whether fragmented public sources can be converted into structured,
time-bounded, provenance-bearing inputs at all. For this feasibility question,
the pipeline was a resource-efficient design: it could retrieve,
normalize, interpret, and trace heterogeneous evidence before a large
task-specific training corpus existed. Nothing prevents the resulting public
data from being used to train a model. Indeed, Part~One's individually proposed feature
candidates and Part~Two's overlapping enumerated pattern candidates reveal the
limits of manual and grammar-based search. Once sufficiently large,
well-covered data exist, learned models become an efficient design choice for
discovering representations, higher-order interactions, and candidate rankings.
The present contribution is upstream and complementary to that next step.

\textbf{One lifecycle, two analytical tasks.} The shared outcome ontology and
evidence-governance rules connect the two parts, but their inferential roles
differ. Part~One asks a pre-founding predictive question using information bounded
to the decision time. Part~Two reconstructs post-founding histories for
retrospective association and behavior analysis; it is not another prediction
task. Combining them demonstrates how the same public-evidence discipline can
support different founder decisions without treating their estimands as
interchangeable.

\textbf{Decision assistance, not automated prescription.} The outputs are
structured prompts for judgment. Part~One's \texttt{PASS}/\allowbreak\texttt{WARN}/\allowbreak%
\texttt{FAIL} verdicts evaluate proposals under an explicit evidence rule;
Part~Two's \texttt{RECOMMEND}/\texttt{AVOID} labels classify observed operating-action
patterns, not companies or investors. This presentation follows the broader
human--AI interaction principle that assistance should expose system capability
and support user control rather than silently replace judgment
\cite{amershi2019guidelines}. Neither vocabulary overrides a founder's
objectives, private information, risk tolerance, or responsibility for a
decision.

\subsection{Limitations and Future Work}
\label{sec:general-limitations}

These results establish feasibility and produce reusable artifacts, but not causal
effects. Part~One also lacks deployment-grade predictive validity, while Part~Two is
retrospective rather than predictive. The main constraints are consolidated in
Table~\ref{tab:limitations}.

\begin{center}
\begin{minipage}{\textwidth}
\centering
\captionof{table}{Principal limitations of the two-part toolchain.}
\label{tab:limitations}
\small
\renewcommand{\arraystretch}{1.15}
\begin{tabular}{>{\raggedright\arraybackslash}p{0.18\textwidth}>{\raggedright\arraybackslash}p{0.74\textwidth}}
\toprule
Component & Principal limitation \\
\midrule
Idea-stage benchmark & Row-disjoint 198-row validation and later 4-fold
out-of-sample tests were completed. From a separately drawn 4{,}000-row
scale-validation cohort, post-stratifying all 1{,}027 completed cases to the
fixed execution-split strata produces $F_{0.5}=0.6506$ [0.598, 0.707], while a
377-row composition-matched subset produces 0.6573 [0.556, 0.746]. Both overlap
the combined benchmark estimate. Post-stratification assumes within-stratum completion
exchangeability; the non-random completion process still limits this independent
holdout evidence; the 1{,}199-row reserved split remains untouched. No tested extension produced a
separated improvement over Full Pipeline v1.5a. The labels measure funding/exit
milestones rather than intrinsic venture quality, and year-constrained retrieval
does not guarantee point-in-time web availability. \\
Vendor case study & Vendor cards combine the June 2026 MIT-licensed OpenSporks
Crunchbase free database snapshot downloaded from Hugging Face with
then-public sources; prices and product claims are time-sensitive,
disclosure depth differs by vendor, estimates are not vendor-confirmed, and the
underlying web screenshots are not distributed. \\
Investor graph & Free-source coverage is uneven: sovereign-wealth coverage is
about 50\%; free-source retrieval captures about 70\% of the interface
evidence visible in the paid-source reference; and per-investor amounts are populated for about 16\% of rows in the separate
full-universe \texttt{funding\_events\_v0.2} table; that field is unpopulated in
the released chip-subset funding table. An earlier cache
accident destroyed 1{,}341 intermediate rows; the full scope was subsequently
recollected, and the current tables supersede that baseline. \\
Event-party attribution & A post-release review of stored titles and
summaries identified self-links in fewer than 1\% of all released event
records: the company and investor endpoints resolve
to the same named entity, although the descriptions
do not substantiate that self-relationship. These include investments in other
companies and fund-level activities; a company legitimately acting as an investor
is not itself an error. This review establishes an internal attribution
limitation, not independent verification of the underlying sources. The released
tables and reported estimates remain unchanged pending source-level adjudication
and reassessment of the derived statistics. \\
Pattern analysis & The registry assigns labels from unadjusted absolute risk differences on a
501-company evaluable chip-startup subset. Outcome-dependent date coverage,
residual confounding, and enumeration over 23{,}308 patterns preclude causal or
broad population claims. A post hoc sensitivity check retains 6{,}313 tests under
Benjamini--Hochberg and 327 under Bonferroni. These findings characterize the
observed chip-company scope. Cross-industry portability is not empirically
tested in this paper; only
the frozen interpretable shortlist is substantive. \\
\bottomrule
\end{tabular}
\end{minipage}
\end{center}

The immediate empirical priority is to complete the separately drawn scale run
for the idea-stage pipeline and then evaluate the frozen pipeline once on the
untouched validation split, without using that split for further tuning. This
should be followed by sensitivity analysis across sectors and founding cohorts. For
the investor graph, the priority is to expand dated failure coverage and improve
observable coverage of governance and operating-support events without conflating
non-observation with inactivity. Future work should also establish a stratified
gold-standard audit of EventChain quality and coverage. This audit should measure
event-type, date, source, and entity-resolution accuracy, explicitly checking
transaction-party roles and company-scope eligibility; inter-reviewer
agreement; and coverage by event class, source type, company outcome, and calendar
period. Comparison with independently curated or higher-coverage reference records
would help distinguish extraction error from events absent from the observable
public record. A separate extension is to apply the same
event-chain method to additional industry cohorts and compare which associations
are domain-specific and which appear transferable, with multiplicity controlled
within each application. This cross-domain evaluation is future work, not an
empirical claim of the present study. Better coverage would also permit
stratified estimates by investor type and stage rather than relying on small
confirmed-action samples.

\subsection{Conclusion}

This paper joins two decisions that are usually studied separately: whether an
idea-stage business model merits pursuit, and which operating actions and capital
partners fit after founding. At the idea stage, Full Pipeline v1.5a yields
$F_{0.5}=0.5357$ [0.412, 0.655] on an independently drawn, row-disjoint
198-row validation sample; the combined 396-row development-plus-validation
estimate is 0.6301, compared with 0.2734 for an apples-to-apples Raw LLM
baseline. In the separately drawn scale cohort, the post-stratified estimate
from all 1{,}027 completed cases is 0.6506 [0.598, 0.707], and the 377-row
composition-matched sensitivity estimate is 0.6573. These are end-to-end system
results rather than component-level attribution. The accompanying AI-inference
case study identifies distribution-layer businesses as the most replicable path
to independent profitability, albeit with a limited revenue ceiling, while
frontier-model ownership offers greater capital-market upside at exceptional
capital cost.

After founding, the provenance-preserving event graph supports both investor
profiles and a label-blind catalog of event-chain associations within completed
chains. In the chip-company implementation, sustained product, customer, and
supply-chain progress is associated with better observed outcomes; financing
participation alone does not establish continuing operating progress.
Two results also qualify intuitive readings: broad investor participation at a
chain's terminal event co-occurs with success even without recorded intermediate actions,
and major-holder changes are negatively associated with outcomes only in the
observed low-follow-through contexts rather than being uniformly adverse.
Funding participation constitutes 79\% of confirmed publicly visible
post-investment actions. The observed evidence tentatively favors considering
strategic corporate investors with relevant acquisition histories for
acquisition-oriented founders, and financing-led institutional VCs with fewer
observable control events for founders prioritizing independent operation.
These findings describe the observed public record; they neither establish
causal effects nor rank companies or investors.

The common contribution is an auditable workflow for turning freely available
public evidence into structured, reusable decision support. The released
ontology, provenance-bearing EventChain data, schemas, benchmarks, and
executable skills form a coherent open toolchain rather than a universal
startup-outcome model. Founders and analysts can inspect the evidence path,
reproduce the data transformations, and treat each verdict or pattern as a
decision aid whose uncertainty and observation boundary remain visible. The
toolchain complements professional venture databases and supports judgment; it
does not replace either professional data infrastructure or the founder's
responsibility for a decision.

\bookmarksetupnext{startatroot}
\section{Data Availability and License}
\label{sec:data}

\begin{sloppypar}
The Part~Two tables are released as \emph{EventChain} on Hugging Face
(\href{https://huggingface.co/datasets/quge007/eventchain}%
{quge007/eventchain}, CC-BY-4.0). The dataset's \texttt{reproduce\_core\_results.py} script
reconstructs the 501-company evaluable cohort and primary P2/P3/P8 counts from
the identity-patched release. The exhaustive 23{,}308-pattern registry is not
included in the dataset release. Executable workflows for both parts are
released as reusable skills in
\href{https://github.com/quge009/startup-decision-skills}%
{quge009/startup-decision-skills} under the MIT license.
\end{sloppypar}

These materials support \emph{procedural reproduction} of the published
workflows and analyses. Deterministic transformations are exactly reproducible
on the same frozen inputs; fresh retrieval and LLM runs need not yield identical
artifacts or results~\cite{blackwell2024reproducible}. New collection should retain
its own retrieval date and provenance.

The Part~One cohort and Part~Two category matching use the MIT-licensed
OpenSporks Crunchbase free snapshot downloaded from Hugging Face
\cite{opensporks_crunchbase}. A conservative redistribution boundary excludes
raw mirror records: Part~One releases aggregate results and methodology;
Part~Two retains matched category tags with source labels. Other public-source
data~\cite{sec_edgar,wikidata} are released as derived analytical fields rather
than raw third-party rows or paid-database exports. Vendor cards and their
internal method, evidence, and link-audit records remain working research
materials, not separate public deliverables.

\section*{Declarations}
\textbf{Funding.} This research received no external funding.

\textbf{Competing interests.} The author declares no competing interests.

\textbf{Author contributions.} Lei Qu designed the study, developed the methods
and software, curated and analyzed the data, and wrote and revised the manuscript.

\textbf{Ethics and responsible use.} The study analyzes public institutional and
company records and involved no intervention with human participants or use of
private personal data. Its outputs are observational decision aids, not investment
advice; users should verify source records and account for entity-resolution and
coverage error.

\section*{Acknowledgements}
This manuscript was drafted with AI-assisted writing tools. The author assumes
responsibility for the analysis, interpretation, source verification, and final
text.

\bookmarksetupnext{startatroot}
\begingroup\sloppy
\bibliographystyle{ACM-Reference-Format}   
\bibliography{references}
\endgroup

\clearpage
\appendix
\bookmarksetupnext{startatroot}
\ifdefined\TenPtClose
\section{Appendix A: Illustrative Candidate Card}
\else
\section{Illustrative Candidate Card}
\fi
\label{app:candidate-card}

This fictional candidate card follows the released candidate-profiler
Markdown format (Section~\ref{sec:p1m-card}). It illustrates the pipeline
input, not an evaluated company or a prediction result.

\begin{Verbatim}
# Candidate card — Workshop Scheduler

**Slug**: workshop-scheduler-example
**Public/private**: private (proposal stage)
**Founding year**: 2025
**Card last updated**: 2026-09-11
**Card author**: hand-authored fictional example
**Original proposal description**: Scheduling software for independent bicycle-repair shops. Charge $49 per shop monthly, plus onboarding. Sell directly to owners; use cloud hosting. Service histories should encourage retention. No customers yet.
**Closest existing vendor analogues**: not assessed; fictional example.

---

## 1. Revenue model

**Primary revenue stream**: subscription.
**Secondary revenue streams**: onboarding fee.
**Pricing structure**:
- Subscription: $49/shop/month (proposed).
- Onboarding: price not disclosed.
**Proposal claim**: "Charge $49 per shop monthly, plus onboarding."
**Plausibility flag**: medium — clear billing unit; demand untested.
**Notes**: Pricing is hypothetical. Contract duration, discounts, conversion, and retention are unknown; no revenue is observed.

---

## 2. Customer segmentation

**Identifiable segments**: independent repair shops (SMBs).
**Estimated mix**:
- Owner-operated shops: share unknown; confidence low.
**Notable disclosed customer logos**: none.
**Proposal claim**: "Sell directly to owners."
**Notes**: Owners are buyers; staff are users. Geography, concentration, and demand from chains remain unspecified. Targeting a segment does not establish adoption.

---

## 3. Cost structure (best-effort estimate)

**GPU spend as % of revenue**: not applicable.
**R&D split**:
- Model training / pre-training: not applicable.
- Inference optimization: not applicable.
- Platform / infra engineering: share unknown.
**S&M as % of revenue**: not disclosed.
**Estimated gross margin**: not disclosed.
**Source of estimate**: proposal-based qualitative inference.
**Notes**: Development, hosting, support, and sales are likely costs. No measured expenses or cost shares are available.

---

## 4. Differentiation / moat

**Differentiation claims**:
- Retention through stored service histories.
**Assessment**:
- Switching costs — mixed: plausible but unverified.
**Notes**: Migration effort, data portability, and competing products need checking. Convenience alone is not a durable moat. No network effect or proprietary technology is established; this hypothesis is input to U-check, not its verdict.

---

## 5. Strategic vulnerabilities

**Identified risks**:
- Competition: alternatives could limit pricing; likelihood unknown, high potential impact.
- Sales costs: acquisition could exceed subscription contribution; likelihood unknown, high potential impact.
**Common categories to consider**:
- Competition; unit-economics; retention.
**Notes**: Easy data export could also weaken retention. These are prospective risks, not observed failures; their relative likelihood is unknown.
\end{Verbatim}

\end{document}